\documentclass[journal]{IEEEtran}

\usepackage{amsmath}
\usepackage{amsfonts}
\usepackage{amssymb}
\usepackage{bm}
\usepackage{isomath}
\usepackage{MnSymbol}
\usepackage{upgreek}
\usepackage{nth}
\usepackage{siunitx}
\usepackage{textcomp}
\usepackage{pifont}
\newcommand{\cmark}{\ding{51}} 
\newcommand{\xmark}{\ding{55}} 

\DeclareMathOperator*{\argmin}{arg\,min}

\usepackage[caption=false,font=normalsize,labelfont=sf,textfont=sf]{subfig}
\usepackage{graphicx}
\usepackage[table]{xcolor} 
\usepackage{svg}

\usepackage{lipsum}

\usepackage{multirow}
\usepackage{tabularx}
\usepackage{booktabs}

\usepackage[nolist,nohyperlinks,printonlyused]{acronym}

\usepackage{float}
\usepackage{stfloats}

\usepackage{url}
\usepackage[numbers,sort&compress]{natbib}
\usepackage[hidelinks]{hyperref}
\hypersetup{
    colorlinks=true,     
    linkcolor=blue,      
    urlcolor=blue,       
    citecolor=blue,      
}
\usepackage{cleveref}
\usepackage{tablefootnote}
\usepackage{orcidlink}

\newcolumntype{C}[1]{>{\centering\let\newline\\\arraybackslash\hspace{0pt}}m{#1}}
\newcolumntype{R}[1]{>{\raggedleft\let\newline\\\arraybackslash\hspace{0pt}}m{#1}}
\newcolumntype{L}[1]{>{\raggedright\let\newline\\\arraybackslash\hspace{0pt}}m{#1}}

\newcommand{\revision}[1]{\textcolor{blue}{#1}} 
\renewcommand{\revision}[1]{#1} 

\title{Continuous-Time State Estimation Methods in Robotics: A Survey}
\author{William Talbot$^{1}$\orcidlink{0009-0008-9426-3595}, Julian Nubert$^{1,2}$\orcidlink{0000-0001-8949-6134}, Turcan Tuna$^{1}$\orcidlink{0000-0001-8662-4890}, Cesar Cadena$^{1}$\orcidlink{0000-0002-2972-6011}, Frederike D\"umbgen$^{3}$\orcidlink{0000-0002-7258-9753}, \newline Jesus Tordesillas$^{4}$\orcidlink{0000-0001-6848-4070}, Timothy D. Barfoot$^{5}$\orcidlink{0000-0003-3899-631X}, Marco Hutter$^{1}$\orcidlink{0000-0002-4285-4990}
\thanks{Manuscript received Month DD, YYYY. Corresponding author: William Talbot.
$^1$Authors are with the Robotic Systems Lab (RSL), ETH Z\"urich, Switzerland.
$^2$Author is with the Max Planck Institute (MPI) for Intelligent Systems, Stuttgart, Germany.
$^3$Author is with Willow, Inria, and the Computer Science Department of ENS, PSL Research University, Paris, France.
$^4$Author is with the Institute for Research in Technology, ICAI School of Engineering, Comillas Pontifical University, Spain.
$^5$Author is with the Autonomous Space Robotics Laboratory (ASRL), University of Toronto, Canada.}
}

\begin{document}

\begin{acronym}[RANSAC] 
    \acro{AA}       [AA]        {Angle-Axis}
    \acro{API}      [API]       {Application Programming Interface}
    \acro{ASRL}     [ASRL]      {Autonomous Space Robotics Lab}
    \acro{BLR}      [BLR]       {Bayesian Linear Regression}
    \acro{BNN}      [BNN]       {Bayesian Neural Network}
    \acro{CLAMP}    [CLAMP]     {Combined Learning from demonstration and Motion Planning} 
    \acro{CT}       [CT]        {Continuous-Time}
    \acro{DT}       [DT]        {Discrete-Time}
    \acro{CAD}      [CAD]       {Computer-Aided Design}
    \acro{CGR}      [CGR]       {Cayley-Gibbs-Rodrigues}
    \acro{DAG}      [DAG]       {Directed Acyclic Graph}
    \acro{DOF}      [DoF]       {Degrees of Freedom}
    \acro{DSO}      [DSO]       {Direct Sparse Odometry}
    \acro{EKF}      [EKF]       {Extended Kalman Filter}
    \acro{EM}       [EM]        {Expectation Maximization}
    \acro{ESGVI}    [ESGVI]     {Exactly Sparse Gaussian Variational Inference}
    \acro{ETH}		[ETH]		{Eidgen\"ossische Technische Hochschule} 
    \acro{FGO}	    [FGO]		{Factor Graph Optimization}
    \acro{FOV}	    [FOV]		{Field of View}
    \acro{FMCW}     [FMCW]      {Frequency-Modulated Continuous-Wave}
    \acro{F2F}	    [F2F]		{Frame-to-Frame}
    \acro{GBP}      [GBP]       {Gaussian Belief Propagation}
    \acro{GLERP}	[GLERP]		{General Linear Interpolation}
    \acro{GMM}      [GMM]       {Gaussian Mixture Model}
    \acro{GN}       [GN]        {Gauss-Newton}
    \acro{GNSS}     [GNSS]      {Global Navigation Satellite System}
    \acro{GP}       [GP]        {Gaussian Process}
    \acrodefplural{GP}[GPs]     {Gaussian Processes}
    \acro{GPDF}     [GPDF]      {Gaussian Process Distance Field}
    \acro{GPGM}     [GPGM]      {Gaussian Process Gradient Map}
    \acro{GPGN}     [GPGN]      {Gaussian Process Gauss-Newton}
    \acro{GPIS}     [GPIS]      {Gaussian Process Implicit Surface}
    \acro{GPM}      [GPM]       {Gaussian Preintegrated Measurement}
    \acro{GPMP}     [GPMP]      {Gaussian Process Motion Planner}
    \acro{GPOM}     [GPOM]      {Gaussian Process Occupancy Map}
    \acro{GPR}      [GPR]       {Gaussian Process Regression}
    \acro{ICP}      [ICP]       {Iterative Closest Point}
    \acro{IF}       [IF]        {Information Filter}
    \acro{IGPMP2}   [iGPMP2]    {Incremental Gaussian Process Motion Planner 2} 
    \acro{IMU}      [IMU]       {Inertial Measurement Unit}
    \acro{ISAM}     [iSAM]     {Incremental Smoothing And Mapping} 
    \acro{ISAM2}    [iSAM2]     {Incremental Smoothing And Mapping 2} 
    \acro{KAN}      [KAN]       {Kolmogorov-Arnold Network}
    \acro{KF}       [KF]        {Kalman Filter}
    \acro{LCM}      [LCM]       {Least Common Multiple}
    \acro{LI}       [LI]        {Linear Interpolation}
    \acro{LIDAR}	[LiDAR]		{Light Detection And Ranging}
    \acro{LIO}      [LIO]       {LiDAR-Inertial Odometry}
    \acro{LLS}      [LLS]       {Linear Least Squares}
    \acro{LM}       [LM]        {Levenberg-Marquardt}
    \acro{LO}       [LO]        {LiDAR Odometry}
    \acro{LPM}      [LPM]       {Linear Preintegrated Measurement}
    \acro{LQR}      [LQR]       {Linear Quadratic Regulator}
    \acro{LTI}      [LTI]       {Linear Time-Invariant}
    \acro{LTV}      [LTV]       {Linear Time-Varying}
    \acro{LVIO}     [LVIO]      {LiDAR-Visual-Inertial Odometry}
    \acro{MAP}      [MAP]       {Maximum A Posteriori}
    \acro{MC}       [MC]        {Monte Carlo}
    \acro{MCMC}     [MCMC]      {Markov Chain Monte Carlo}
    \acro{MDC}      [MDC]       {Motion Distortion Correction}
    \acro{MHE}      [MHE]       {Moving Horizon Estimation}
    \acro{MLE}      [MLE]       {Maximum Likelihood Estimate}
    \acro{MLP}      [MLP]       {Multilayer Perception}
    \acro{MVO}      [MVO]       {Multimotion Visual Odometry}
    \acro{NLLS}     [NLLS]      {Nonlinear Least Squares}
    \acro{NTV}      [NTV]       {Nonlinear Time-Varying}
    \acro{NURBS}    [NURBS]     {Non-Uniform Rational B-Spline}
    \acro{ODE}		[ODE]		{Ordinary Differential Equation}
    \acro{PC}       [PC]        {Point Cloud}
    \acro{PDL}      [PDL]       {Powell's Dogleg}
    \acro{PF}       [PF]        {Particle Filter}
    \acro{PTAM}     [PTAM]      {Parallel Tracking And Mapping} 
    \acro{QCQP}		[QCQP]		{Quadratically Constrainted Quadratic Program}
    \acro{QLB}		[QLB]		{Quaternion Linear Blending}
    \acro{RANSAC}	[RANSAC]	{RANdom SAmple Consensus}
    \acro{RFF}		[RFF]		{Random Fourier Features}
    \acro{RGBD}     [RGB-D]     {RGB-Depth}
    \acro{RK}       [RK]        {Runge-Kutta}
    \acro{RPE}		[RPE]		{Relative Pose Error}
    \acro{RPY}      [RPY]       {Roll Pitch Yaw}
    \acro{RS}       [RS]        {Rolling Shutter}
    \acro{RSC}      [RSC]       {Rolling Shutter Compensation}
    \acro{RSL}		[RSL]		{Robotics Systems Laboratory}
    \acro{RMSE}		[RMSE]		{Root Mean Squared Error}
    \acro{RVM}      [RVM]       {Relevance Vector Machine}
    \acro{SAM}		[SAM]		{Smoothing And Mapping}
    \acro{SD}		[SD]		{Steepest Descent}
    \acro{SDE}		[SDE]		{Stochastic Differential Equation}
    \acro{SDP}		[SDP]		{Semidefinite Program}
    \acro{SFM}		[SfM]		{Structure from Motion}
    \acro{SLAM}		[SLAM]		{Simultaneous Localization And Mapping}
    \acro{SLAP}		[SLAP]		{Simultaneous Localization And Planning}
    \acro{SLERP}	[SLERP]		{Spherical Linear Interpolation}
    \acro{SQP}      [SQP]       {Sequential Quadratic Programming}
    \acro{STEAM}	[STEAM]		{Simultaneous Trajectory Estimation And Mapping}
    \acro{STEAP}	[STEAP]		{Simultaneous Trajectory Estimation And Planning}
    \acro{SVM}      [SVM]       {Support Vector Machine}
    \acro{TBF}		[TBF]		{Temporal Basis Function}
    \acro{TGP}		[TGP]		{Temporal Gaussian Process}
    \acrodefplural{TGP}[TGPs]   {Temporal Gaussian Processes}
    \acro{TE}		[TE]		{Trajectory Estimation}
    \acro{UAV}		[UAV]		{Unmanned Aerial Vehicle}
    \acro{UGPM}     [UGPM]      {Unified Gaussian Preintegrated Measurement}
    \acro{UKF}      [UKF]       {Unscented Kalman Filter}
    \acro{UPM}      [UPM]       {Upsampled Preintegrated Measurement}
    \acro{UWB}      [UWB]       {Ultra-WideBand}
    \acro{VI}       [VI]        {Visual-Inertial}
    \acro{VIO}		[VIO]		{Visual-Inertial Odometry}
    \acro{VO}		[VO]		{Visual Odometry}
    \acro{VTR}		[VT\&R]		{Visual Teach And Repeat}
    \acro{WNOA}		[WNOA]		{White-Noise-On-Acceleration}
    \acro{WNOD}		[WNOD]		{White-Noise-On-Derivative}
    \acro{WNOJ}		[WNOJ]		{White-Noise-On-Jerk}
    \acro{WNOV}		[WNOV]		{White-Noise-On-Velocity}
    \acro{WO}		[WO]		{Wheel Odometry}
\end{acronym}

\maketitle

\begin{abstract}
Accurate, efficient, and robust state estimation is more important than ever in robotics as the variety of platforms and complexity of tasks continue to grow.
Historically, discrete-time filters and smoothers have been the dominant approach, in which the estimated variables are states at discrete sample times.
The paradigm of continuous-time state estimation proposes an alternative strategy by estimating variables that express the state as a continuous function of time, which can be evaluated at any query time.
Not only can this benefit downstream tasks such as planning and control, but it also significantly increases estimator performance and flexibility, as well as reduces sensor preprocessing and interfacing complexity.
Despite this, continuous-time methods remain underutilized, potentially because they are less well-known within robotics.
To remedy this, this work presents a unifying formulation of these methods and the most exhaustive literature review to date, systematically categorizing prior work by methodology, application, state variables, historical context, and theoretical contribution to the field.
By surveying splines and Gaussian processes together and contextualizing works from other research domains, this work identifies and analyzes open problems in continuous-time state estimation and suggests new research directions.

\end{abstract}


\section{Introduction}
\label{sec:introduction}

\begin{figure*}
    \centering
    \includegraphics[trim={15 13 10 15},clip,width=\textwidth]{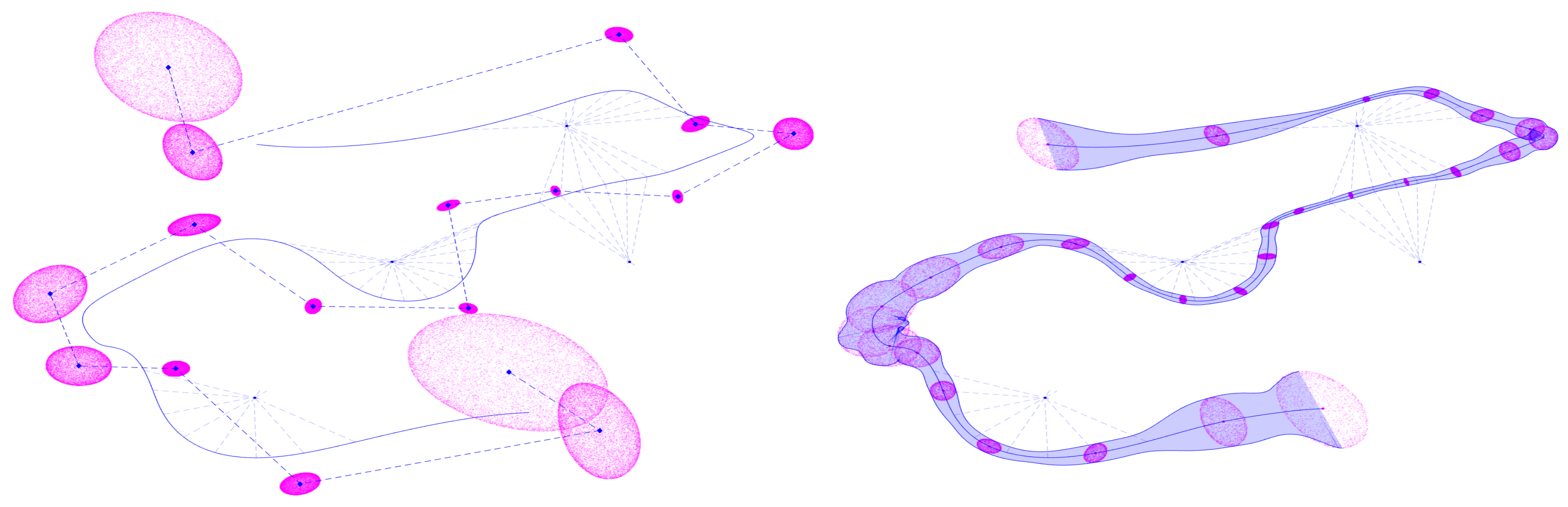}
    \vspace{-6mm}
    \caption{Example of 2D localization from noisy accelerometer, gyroscope, and range-bearing measurements (shown projected from interpolated poses) using two popular \ac{CT} methods; B-splines (left), and `exactly sparse' Gaussian processes (right).
    The Laplace approximation for the posterior (\Cref{eqn:laplace_approximation}) is obtained via batch optimization.
    For the uniform cubic B-spline, the $SE(2)$ control points (diamonds) are estimated, while for the uniform \emph{`constant-jerk'} Gaussian process, $SE(2)$ states on the trajectory are estimated.
    The $3\sigma$ uncertainties for the estimated variables (pink) and the interpolated $SE(2)$ trajectories (blue) are shown.
    Since the latter method supports efficient covariance interpolation in manifold spaces, a $3\sigma$ uncertainty envelope is shown around the trajectory.}
    \vspace{-5mm}
    \label{fig:eye_candy}
\end{figure*}

\IEEEPARstart{A}{utonomous} driving, extra-planetary exploration, infrastructure inspection, and environmental monitoring are examples of the complex tasks expected of robotic platforms today.
To cope with these diverse challenges, onboard sensors have increased in quantity, variety, and bandwidth, and the requirement for accurate and robust state estimation is more critical than ever.
The constraints of these platforms, including cost, size, sensor quality, power consumption, and computational resources, vary significantly.
Some platforms, like commercial drones or handheld devices, are highly constrained, while autonomous cars, construction machines, and others may have access to more onboard compute.
As the demand for such diverse robotic systems grows in the coming decades, so will the requirement for general, flexible, and scalable state estimation solutions.
A paradigm shift, from discrete to continuous-time, may be the key to achieving this.

\paragraph*{\revision{\acf{DT}}}
Historically, state estimation techniques have operated in discrete-time, which means the state of the robot or process is estimated at specific moments in time.
However, the evolution of the system is not modeled, and intermediate states cannot be inferred.
These times must include the measurement times, so regardless of how quickly the system state evolves, \ac{DT} methods require estimating variables at all of these sample times~\cite{lynen2013robust,bloesch2017two,nubert2022graph}.
The computation of such algorithms scales poorly as the number and frequency of sensors increase, compromising their ability to run in real-time.
Several techniques and tricks are employed to work around this issue.
One of the most common is to aggregate measurements over time, especially from sensors that capture hundreds or thousands of measurements per second, to form lower-frequency \emph{pseudo}-measurements.
Examples include the accumulation of points from a sweeping \ac{LIDAR} sensor for inter-scan or scan-to-map registration~\cite{zhang2014loam,zhang2017low,li2019net,nubert2021self,tuna2023x}, feature extraction in event~\cite{vidal2018ultimate} or \ac{RS} cameras, and \ac{IMU} preintegration~\cite{forster2015imu,forster2016manifold,brossard2021associating,le20183d,le2020gaussian}.
Motion distortion may occur if the platform moves substantially over this accumulation period.
\ac{MDC} methods such as \ac{LIDAR} deskewing~\cite{wu2021detailed} or \ac{RS} compensation~\cite{nicklin2007rolling} attempt to correct for this.
However, \ac{MDC} is fundamentally a \emph{chicken-and-egg} problem since the motion-corrected measurements are required to estimate the motion accurately.
Consequentially, these methods introduce hard-to-model errors into the system.
Some \ac{DT} systems (e.g., ~\cite{bahnemann2021under}) also achieve state reduction by triggering sensors to capture simultaneously and often have complicated hardware triggering systems and clock synchronization in place to ensure that the timestamps reported by sensors are well-aligned to a central clock.
Yet, in systems without proper time synchronization, estimating time offsets between sensor clocks is a complex task for \ac{DT} state estimators~\cite{furgale2013unified,cioffi2022continuous}.

\paragraph*{\revision{\acf{CT}}}
Continuous-time methods (\Cref{fig:eye_candy}) differ by modeling the underlying process as a continuous function of time and have the defining characteristic that the state can be queried at any time (sometimes within certain bounds). 
This capability can, on the one hand, be useful for downstream tasks that rely on accurate state estimates, such as planning or control.
On the other hand, it allows querying \emph{interpolated} states at measurement times during estimation without requiring additional explicitly defined variables.
By providing a layer of abstraction between the time-varying state and the optimization variables, \ac{CT} methods allow the number of variables to be flexible and their generation to be application and sensor-agnostic.
This property facilitates an elegant means of trading off accuracy and computation time while considering the robot's dynamics.
\revision{Such a system encourages the direct inclusion of asynchronous high-frequency sensor measurements, such as from an \ac{IMU} or \ac{LIDAR}, without necessarily creating new estimation variables, avoiding the errors introduced by measurement aggregation.
Furthermore, these systems can jointly estimate the time offsets between sensor clocks by shifting the interpolation time~\cite{cioffi2022continuous}.}

\paragraph*{\revision{Significance of this Work}}
The \ac{CT} paradigm could be the key to accurate, robust, and scalable state estimation.
Indeed, \ac{CT} methods are already being leveraged to achieve state-of-the-art performance across a full spectrum of applications.
However, they remain relatively niche and less well-understood within the broader robotics community.
The field has progressed substantially since the last \ac{CT} survey almost a decade ago~\cite{furgale2015continuous}.
Other review works have focused more specifically on a single method~\cite{cioffi2022continuous} or sensor modality~\cite{lee2024lidar}, or on comparative performance~\cite{johnson2024continuous}.
To update the community on modern \ac{CT} methods and lower the bar of entry to future research, this work makes following contributions:
\begin{itemize}
    \item A concise background (\Cref{sec:state_estimation_in_robotics}) and consolidated formulations for the established \ac{CT} methods (\Cref{sec:theory}).
    \item The most complete survey of \ac{CT} state estimation to date (until October 2024), categorized by method, application, state variables, historical context, or theoretical contribution, as most appropriate (\Cref{sec:survey}).
    \item Identification and analysis of open problems (\Cref{sec:open_problems}).
    \item A discussion of the scope of \ac{CT} methods in other robotics' domains and applications (\Cref{sec:applications_to_other_domains}).
\end{itemize}



\section{State Estimation in Robotics}
\label{sec:state_estimation_in_robotics}

\subsection{Preliminaries}
\label{sec:state_estimation_in_robotics:preliminaries}

This section provides a brief introduction to the field of state estimation and establishes a common notation for \ac{DT} estimation and the \ac{CT} methods introduced in \Cref{sec:theory}.
In the interests of conciseness, a detailed theory of optimization-based state estimation theory will not be covered as has been done in prior works~\cite{grisetti2010tutorial,barfoot2017state,dellaert2017factor,dellaert2021factor}.
This work considers state estimation problems which aim to tractably estimate the probability distribution of a (finite) set of (continuous) random variables that parameterize a system of interest.
Lie theory~\cite{gilmore2006lie,chirikjian2011stochastic,sola2018micro} has proven to be a valuable framework for treating these variables accurately in the context of state estimation in robotics;
\revision{e.g.}, the $SO(n)$ and $SE(n)$ groups for representing $n$-dimensional orientations and poses.
Because Lie groups are smooth, differentiable manifolds, they have unique tangent spaces, which are vector spaces amenable to the linear algebra operations required during optimization.
This work will adopt the notation and \emph{right}-handed conventions of \citet{sola2018micro}.

\subsubsection{\revision{States}}
A continuously evolving process is defined by two components; \textit{i)} a time-varying state, denoted as $\mathbf{x}(t)$ for time variable $t$, such as the $SE(3)$ pose of a robot, and \textit{ii)} its time-invariant context, a set of variables denoted as $\bm{\theta}$, such as sensor calibration or map parameters.
Without yet proposing a relationship to $\mathbf{x}(t)$, a set of variables $\mathbf{x} = (\mathbf{x}_0, \dots, \mathbf{x}_N)$ is defined to describe $\mathbf{x}(t)$.
Estimating these variables, collected together as $\mathbf{\Theta} = \{\mathbf{x}, \bm{\theta}\}$, will then be possible \revision{by incorporating} measurements and prior knowledge.

\subsubsection{\revision{Measurements}}
A measurement model describes how an observation $\mathbf{\tilde{z}}$ is generated from one or more sensors.
The \emph{measurement function} for this model,
$\mathbf{z} = \mathbf{h}(\mathbf{x}, \bm{\theta}, \mathbf{n})$,
defines a (measured) observation as a function of the state and associated noise $\mathbf{n}$, usually assumed to be drawn from a Gaussian distribution and independent between measurements.
Common examples in robotics include range-bearing, projective geometry, kinematic, and inertial measurement models.
A \emph{residual function} is defined to obtain a vector-valued residual $\mathbf{e} = \mathbf{g}(\mathbf{z}, \mathbf{\tilde{z}})$, which is commonly subtraction or the generalized $\ominus$ operator~\cite[Equation~(26)]{sola2018micro} for Lie groups.
Alternatively, $\mathbf{h}(\cdot)$ and $\mathbf{g}(\cdot)$ may be implicitly combined such as in point-to-plane~\cite{zhang2017low,zheng2024trajlo} or on-unit-sphere angular~\cite{hug2020hyperslam} distances.

\subsubsection{\revision{Priors}}
Prior models \revision{capture} assumptions about a system before evidence from measurement models is incorporated.
They can be formulated in the same way as measurement models, except that $\mathbf{\tilde{z}}$ is not observed but set \emph{a priori}.
In practice, a belief about the system's initial conditions (e.g., pose and velocity) is often required to ensure a unique solution in the optimization.
A more complex example \revision{is} planar surface regularization \revision{in highly structured environments}.
As with measurement models, residuals can be defined for such priors.

\subsubsection{\revision{Processes}}
Process models (a.k.a. system dynamics, motion models) describe the evolution of the state over time.
Process models have frequently appeared in \ac{DT} filters, occasionally in optimization-based approaches, and are a fundamental idea used in some \ac{CT} methods.
Since the process model is known \emph{a priori} and does not require measurements, it can be seen as a type of Bayesian prior.
If the process is \emph{Markovian}, then it can be formulated in terms of the current process state $\mathbf{x}(t)$, process inputs $\mathbf{u}(t)$, zero-mean process noise $\mathbf{w}(t)$, and context $\bm{\theta}$, as a \ac{NTV} \ac{SDE}
\begin{equation}
    \mathbf{\dot{x}}(t) = \mathbf{f}(\mathbf{x}(t), \bm{\theta}, \mathbf{u}(t), \mathbf{w}(t))
    .
    \label{eqn:pm_full}
\end{equation}
Commonly, $\mathbf{f}$ does not depend on the time-invariant parameters $\bm{\theta}$ and is linearized about a current estimate of the other inputs:
\begin{gather}
    \dot{\mathbf{x}}(t) \approx \mathbf{F}(t)\mathbf{x}(t) + \mathbf{v}(t) + \mathbf{L}(t)\mathbf{w}(t)
    ,
    \label{eqn:ltvsde}\\
    \text{with} ~ \mathbf{v}(t) \coloneq \mathbf{f}(\mathbf{\bar{x}}(t), \mathbf{u}(t), \mathbf{0}) - \mathbf{F}(t) \mathbf{\bar{x}}(t)
    ,
    \label{eqn:v(t)}\\
    \left. \mathbf{F}(t) \coloneq \frac{\partial \mathbf{f}}{\partial \mathbf{x}} \right|_{\mathbf{\bar{x}}(t), \mathbf{u}(t), \mathbf{0}}
    , ~ \text{and} ~ 
    \left. \mathbf{L}(t) \coloneq \frac{\partial \mathbf{f}}{\partial \mathbf{w}} \right|_{\mathbf{\bar{x}}(t), \mathbf{u}(t), \mathbf{0}}
    .
    \label{eqn:F(t)_L(t)}
\end{gather}
The general solution of this \ac{LTV} \ac{SDE}~\cite{maybeck1982stochastic,stengel1994optimal}, given some initial state $\mathbf{x}(t')$ at time $t'$, is
\begin{equation}
    \resizebox{0.90\linewidth}{!}{$\displaystyle
    \mathbf{x}(t) = \mathbf{\Phi}(t, t') \mathbf{x}(t') + \int_{t'}^t \mathbf{\Phi}(t, s) \left(\mathbf{v}(s) + \mathbf{L}(s)\mathbf{w}(s)\right) ds
    $}
    .
\label{eqn:ltvsde:solution}
\end{equation}
\normalsize
Here, $\mathbf{\Phi}(t, t')$ is the \emph{transition matrix} for the process, with properties $\mathbf{\Phi}(t, t') = \mathbf{\Phi}(t, s)\mathbf{\Phi}(s, t')$ for any $s \in [t', t]$, and $\mathbf{\Phi}(t, t) = \mathbf{I}$.
It can be found analytically for simple \acs{LTV}-\acsp{SDE} but can be challenging to find in general, and so without an analytical solution, it must be computed numerically~\cite{barfoot2014batch,anderson2015batch,anderson2017batch}.

\subsubsection{\revision{\texorpdfstring{\ac{MAP}}{} Estimation}}
Modern state estimation approaches are fundamentally probabilistic, relying on the principle of Bayesian inference.
The application of Bayesian theory to \ac{DT} state estimation in robotics has been covered thoroughly in prior work~\cite{dellaert2017factor,dellaert2021factor,thrun2002probabilistic,sarkka2023bayesian}, and the findings transfer to \revision{continuous-time}.
One powerful consequence of this theory is that under the assumption that all residual errors (e.g., from measurement, prior, and process models) are drawn from zero-mean Gaussian distributions, an optimal estimate (in the \ac{MAP} sense) of the state is obtainable by minimizing the sum of their squared Mahalanobis distances.
Specifically, for $M$ residuals with individual covariances $\mathbf{\Sigma}_j$, a Gaussian approximation for the posterior distribution $\mathcal{N}(\mathbf{\Theta}_\text{MAP}, \mathbf{\Sigma}_{\text{MAP}})$ (a.k.a. \emph{Laplace approximation}) is found~\cite{rosen2014inference} by solving
\begin{align}
    \begin{split}
        \mathbf{\Theta}_{\text{MAP}} &:= \argmin_{\mathbf{\Theta}} \sum_{j = 0}^{M} \lVert \mathbf{e}_j \rVert_{\mathbf{\Sigma}_j}^2
        ,\\
        \mathbf{\Sigma}_{\text{MAP}}^{-1} &:= {\left. \frac{\partial^2}{\partial \mathbf{\Theta}^2} \sum_{j = 0}^{M} \lVert \mathbf{e}_j \rVert_{\mathbf{\Sigma}_j}^2 \right|_{\mathbf{\Theta}_{\text{MAP}}}}
        .
    \end{split}
    \label{eqn:laplace_approximation}
\end{align}

\subsection{\acf{DT} State Estimation}
\label{sec:state_estimation_in_robotics:discrete_time_state_estimation}
In \acf{DT} state estimation, the process state variables $\mathbf{x}$ to be estimated are defined as the process state itself at a specific set of times $(t_0, \dots, t_N)$, i.e., $\mathbf{x}_i \coloneq \mathbf{x}(t_i)$.
There are two broad categories of approaches for solving \ac{DT} state estimation problems: \emph{filtering} and \emph{smoothing}.

\subsubsection{\revision{Filtering}}
Bayesian filtering adopts a predict \& update pattern that relies on the Markov assumption.
This assumption states that the evolution of a process, such as robot motion, can be inferred from its current state only and does not depend on how this state was reached.
In the \textit{predict} step, the prior belief of the state is propagated according to a process model.
Then, in the \textit{update} step, measurements are used to estimate the posterior distribution for the state.
\acfp{KF}~\cite{kalman1960new,julier1997new,wan2000unscented} and \acfp{IF}~\cite{thrun2002probabilistic} have proven to be highly effective and efficient, and are still widely used, especially when latency and computational load must be low.
If the state has a non-Gaussian distribution, discretized state-space methods like the histogram filter, or \acl{MC} methods such as the \acf{PF}, which uses samples to represent the state probability distribution, are often used instead~\cite{thrun2002probabilistic}.

\subsubsection{\revision{Smoothing}}
Bayesian smoothing estimates the joint probability distribution of variables representing (a window of) the current and previous robot states, typically relying on optimization techniques.
Smoothing methods have gained popularity in many robotics problems as they have matured in recent decades regarding robustness, accuracy, and speed.
Due to the Markov property, variables involved in the optimization are usually related to only a small subset of (neighboring) variables.
These approaches demonstrate speeds compatible with onboard robot operation by exploiting this inherent (near diagonal) sparsity through efficient linear algebra tools (e.g., sparse LU, LDL, Cholesky, and QR factorization), which are the basis of efficient optimal solutions of linear problems and fast iterative solvers based on relinearization for nonlinear problems (e.g., \acf{GN}~\cite{barfoot2017state}, \acl{LM}, \acl{SD}, \acl{PDL})~\cite{dellaert2017factor}.

\subsection{\acf{CT} State Estimation}
\label{sec:state_estimation_in_robotics:continuous_time_state_estimation}

Since processes in the physical world usually change continuously over time, it is natural to seek a continuous-time representation to model them.
The fundamental property of \ac{CT} methods is their ability to query the state $\mathbf{x}(t)$ at any required time $t$, given a set of variables $\mathbf{x} = (\mathbf{x}_0, \dots, \mathbf{x}_N)$.
The exact relationship between $\mathbf{x}(t)$ and $\mathbf{x}$ depends on the chosen method. 
Yet, a key advantage of these methods is that the number and temporal placement of variables are flexible.
Intuitively, the quantity of variables must scale with the duration of the process and needs to be increased to accurately model more complex dynamics.
Notably, these methods do not require variables for every measurement time and can naturally accommodate asynchronous sensor data through their inherent ability to interpolate (or extrapolate).
To remain computationally competitive with \ac{DT} methods, \ac{CT} methods rely on the \emph{local support} property, meaning that inferring a state at a particular time depends only on a small number of (local) variables.
\looseness=-1

The robotic research community has experimented with several \ac{CT} methods for state estimation over the past years.
\textbf{\textit{i)}} Interpolation (e.g., \acf{LI}) and integration methods (\Cref{sec:theory:interpolation_and_integration}) appear frequently, in part due to their simplicity and low computational complexity.
\textbf{\textit{ii)}} Temporal splines (\Cref{sec:theory:temporal_splines}), especially B-splines, are among the most popular \ac{CT} methods to emerge.
Through \emph{control points} and \emph{knots}, splines embrace the abstraction of state inference from estimated variables.
\textbf{\textit{iii)}} `Exactly Sparse' \acfp{TGP}\footnote{
These are not to be confused with \emph{sparse} \acsp{GP}~\cite{quinonero2005unifying}, which address the cubic complexity of \acsp{GP} by selecting \emph{active points} as a subset of training points or optimizable pseudo-inputs that account for the entire training set~\cite{ko2009gp}.} (\Cref{sec:theory:temporal_gaussian_processes}) are a recent, compelling approach.
\Cref{tab:continuous_time_method_comparison} provides a high-level overview.
\revision{The spline and \acs{GP} approaches have adopted the optimization-based smoothing paradigm, with few exceptions~\cite{li2023embedding,shen2024ctemlo}, which may be partly attributed to the way in which these methods are conventionally formulated.
Hence, this work focuses primarily on their use in an optimization context.}
\looseness=-1


\begin{table*}[ht]
    \centering
    \caption{Comparison of continuous-time state estimation methods.}
    \vspace{-3mm}
    \resizebox{\textwidth}{!}{
    \def\arraystretch{0.9}
    \rowcolors{2}{gray!8}{white}
    \begin{tabular}{|l|lll|}
        \hline
        \textbf{Specifications} & \textbf{Linear Interpolation} & \textbf{Temporal Splines} & \textbf{\acp{TGP}} \\
        \hline
        Interpolation & linear & polynomial & process model \\
        Lie Group Support & \cmark & \cmark & \cmark \\
        Covariance Interpolation & \cmark & \Cref{sec:open_problems:temporal_splines:covariance_interpolation} & \cmark \\
        Extrapolation & \cmark & \cite{persson2021practical} & \cmark \\
        Non-Interpolated Residual Support & \cmark & \xmark & \cmark \\
        Initialization & linear extrapolation & \Cref{sec:open_problems:temporal_splines:initialization} & process model \\
        \hline
        Process Variables $\mathbf{x}$ & explicit states & control points & explicit (Markovian) states \\
        Derivatives & constant \nth{1} derivative & computable (exact) & within state (probabilistic) \\
        Variable $\mathbf{x}_i$ Size & size of $\mathcal{L}$ & size of $\mathcal{L}$ & (size of $\mathcal{L}$) + (\acs{DOF} of $\mathcal{L}$) $\times$ (derivatives) \\
        Variables per Interpolation & 2 & $k$ & 2 \\
        \hline
        \cellcolor{white} & - & spline type (e.g., B-spline) & process model (e.g., \acs{WNOA}) \\
        \cellcolor{white} & state representation (e.g., joint/split pose) & state representation (e.g., joint/split pose) & state representation (e.g., joint/split pose) \\
        \cellcolor{white} & - & order $k$ (e.g., $k=4$ for cubic) & hyperparameters (e.g., $\mathbf{Q}_C$) \\
        \cellcolor{white}\multirow{-4}{*}{\centering \parbox{4cm}{\vspace*{-1.5cm} Design Choices \vspace*{-1.5cm}}} & state times & knots & state times \\
        \hline
        \cellcolor{white} & parametric & parametric & non-parametric \\
        \cellcolor{white}\multirow{-2}{*}{\centering \parbox{4cm}{\vspace*{-1.5cm} Characteristics \vspace*{-1.5cm}}} & 2 \acsp{TBF} & $k$ \acsp{TBF} & $\infty$ \acsp{TBF} (kernel trick) \\
        \hline
    \end{tabular}
    }
    \vspace{-5.5mm}
    \label{tab:continuous_time_method_comparison}
\end{table*}

\section{Theory}
\label{sec:theory}

\subsection{Interpolation \& Integration}
\label{sec:theory:interpolation_and_integration}

When interpolation and integration are discussed in the context of \ac{CT} state estimation, the state variables precisely represent the process state (i.e., $\mathbf{x}_i \coloneq \mathbf{x}(t_i)$).
Interpolation is defined as the inference of intermediate states from estimated curves that pass exactly through these explicit states.\footnote{
However, the term interpolation is frequently overloaded, and so is used more liberally in later sections to describe inference with regression models that do not pass through every data point.
This is common in the spline literature, although only some spline types are \emph{interpolating}.}
On the other hand, integration uses rate measurements, such as the angular rate from an \ac{IMU}, for this inference.

\subsubsection{\acf{LI}}
\label{sec:theory:interpolation_and_integration:linear_interpolation}

One of the simplest and fastest interpolation methods is \acf{LI}.
The \textit{generalized} \ac{LI} of two adjacent states can be written as
\begin{equation}
    \text{GLERP}(\mathbf{x}_i, \mathbf{x}_{i+1}, \alpha) = \mathbf{x}_i \boxplus (\alpha \cdot (\mathbf{x}_{i+1} \boxminus \mathbf{x}_i))
    ,
    \label{eqn:glerp}
\end{equation}
where $\boxplus, \boxminus$ are the addition and subtraction operators for composite manifolds~\cite[Section IV]{sola2018micro}, and $\alpha := \frac{t - t_i}{t_{i+1} - t_i} \in [0, 1)$ for interpolation time $t \in [t_i, t_{i+1})$ (outside is extrapolation).
For Lie groups, these operators become $\oplus, \ominus$~\cite[Equations~(25) and (26)]{sola2018micro}, which when applied to quaternions~\cite{haarbach2018survey} or rotation matrices~\cite{barfoot2017state} yield \emph{spherical linear interpolation} (\acs{SLERP}).
This has also been applied to $SE(3)$ poses~\cite{barfoot2017state}.
When the states belong to vector space, this simplifies to
\begin{equation}
    \resizebox{0.89\linewidth}{!}{$\displaystyle
    \text{LERP}(\mathbf{x}_i, \mathbf{x}_{i+1}, \alpha) := \mathbf{x}_i + \alpha (\mathbf{x}_{i+1} - \mathbf{x}_i) = (1 - \alpha) \mathbf{x}_i + \alpha \mathbf{x}_{i+1}
    $}
    .
    \label{eqn:lerp}
\end{equation}
Importantly, \ac{LI} maintains a constant rate of change (`velocity') while other interpolation methods, such as \ac{QLB}~\cite{haarbach2018survey} do not.
\ac{LI} therefore enables rapid inference of the state at any time under this constant-rate assumption, and allows the number of optimized states to be chosen flexibly.
While nonlinear interpolation methods exist, they are usually captured within the scope of \acfp{TBF}, the foundation for temporal splines~\cite{furgale2015continuous} and \acp{TGP}~\cite{tong2013gaussian}.
Indeed, \ac{LI} is exactly a special case of these methods.
Despite the development of more advanced methods, \ac{LI} remains popular for its simplicity and speed, and may be sufficiently accurate to model many processes.

\subsubsection{Numerical Integration}
\label{sec:theory:interpolation_and_integration:numerical_integration}

Numerical integration in robotics is sometimes discussed in the \ac{CT} state estimation literature, as it allows inference of intermediate states from rate measurements.
However, unlike the other approaches, numerical integration does not provide a closed-form state inference function.
In robotics, this is often used for the motion compensation of scanning sensors such as \acsp{LIDAR} or \ac{RS} cameras, or for \emph{\acs{IMU} preintegration}.
A common method is \emph{Riemann summation} (a.k.a. the \emph{rectangle rule}), which assumes a constant rate of change and hence is closely related to \ac{LI}.
When formulated as an \acs{ODE}, numerically integrating constant rates over short time intervals is known as the \emph{forward Euler method}.
Regressing and integrating \acp{GP} for \ac{IMU} preintegration has also been explored (\Cref{sec:survey:interpolation_and_integration:gp_imu_preintegration}).

\subsection{Temporal Splines}
\label{sec:theory:temporal_splines}

A number of different formulations for splines have been presented in the literature~\cite{hug2020hyperslam,haarbach2018survey,furgale2015continuous,anderson2014hierarchical,de1972calculating,de1978practical,crouch1999casteljau,qin1998general,sommer2020efficient,persson2021practical}.
The formulation presented in this work is primarily based on the works of \citet{furgale2015continuous} and \citet{sommer2020efficient}.
A \emph{spline} is a function of a scalar input whose resultant curve (in vector or manifold space) is a piecewise polynomial of \emph{degree} $m$.\footnote{
Technically, splines are a special case of \acfp{TBF} where the $k$ bases are the polynomials of degree 0 to $m$.}
The generated curve is composed of \emph{segments}.
The \emph{order} $k \in \mathbb{Z}^{+}$ is a fundamental property of a spline guaranteeing certain continuity properties and is related to the degree as $k = m + 1$.
In \emph{temporal} splines, the input scalar values represent time.

A set of \emph{control points} $\mathbf{x} = (\mathbf{x}_0, \mathbf{x}_1, \dots)$ (the estimation variables) and a set of scalar times called \emph{knots} $(t_0, t_1, \dots)$ characterize a temporal spline. 
Each spline segment depends only on $k$ control points.
This is the \emph{local support} property for splines and is key for efficient computation and optimization.
The $i$-th polynomial segment (indexing from $i=0$) depends on a subset of the control points $(\mathbf{x}_i, \mathbf{x}_{i+1}, \dots, \mathbf{x}_{i + k - 1})$ and is defined on the interval $[t_{i + k - 2}, t_{i + k - 1})$.\footnote{
To capture the degenerate $k = 1$ case in a manner consistent with higher orders, define $t_{-1} \coloneq -\infty$, and when no knots are defined, $t_{0} \coloneq \infty$.}
A time $t$ lying within this interval can be normalized as
\begin{equation}
    u_i(t) = \frac{t - t_{i + k - 2}}{t_{i + k - 1} - t_{i + k - 2}} \in [0, 1)
    .
    \label{eqn:spline_ui}
\end{equation}
Polynomial powers of this normalized time are then stacked into the vector 
$\mathbf{u}_i(t) = \begin{bmatrix}1 & u_i(t) & u_i(t)^2 & \dots & u_i(t)^{m}\end{bmatrix}^T$.

\subsubsection{Interpolation}
\label{sec:theory:temporal_splines:interpolation}

\begin{figure}[t]
    \centering
    \includegraphics[width=\linewidth,trim={145, 65, 85, 25},clip]{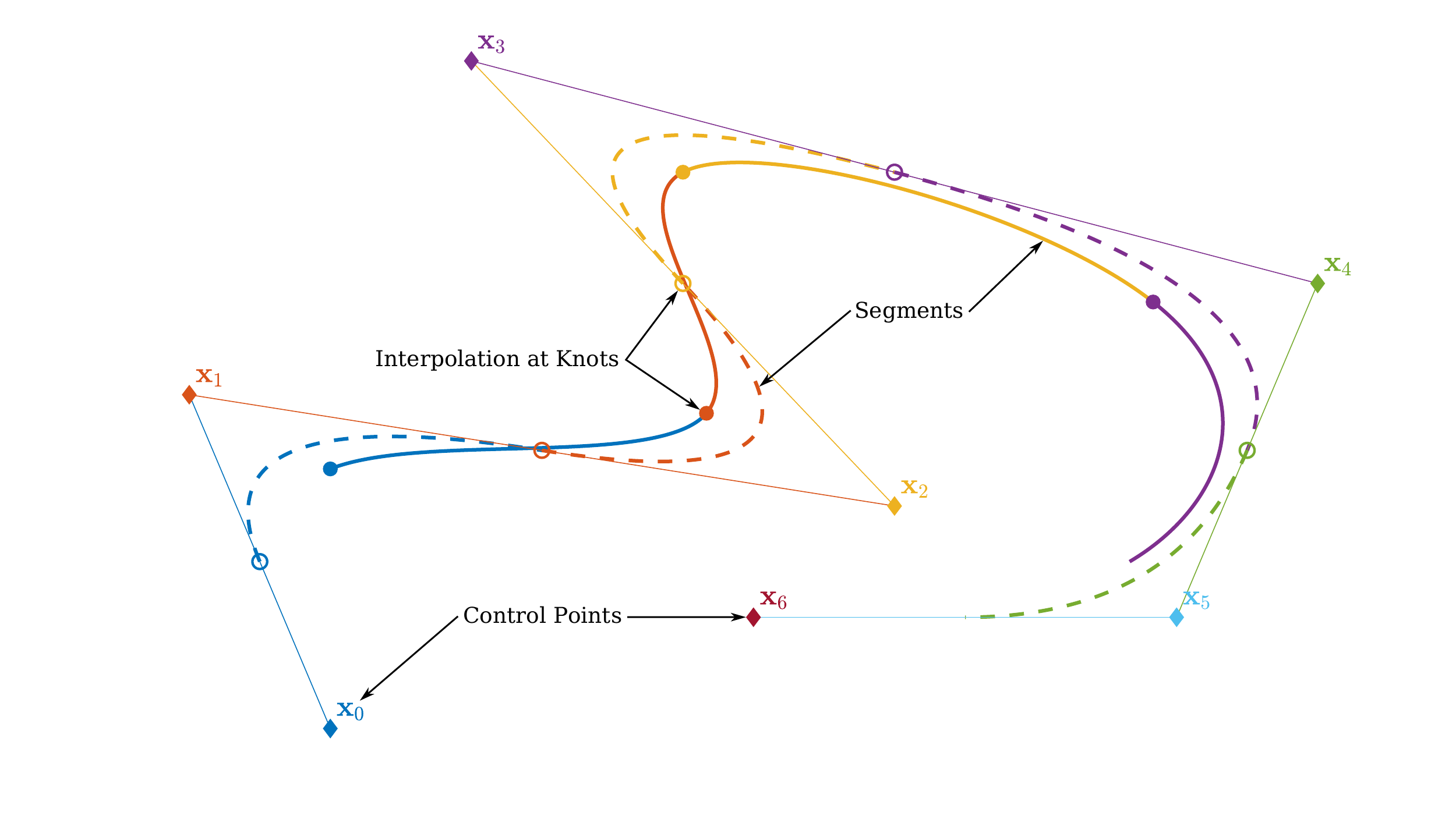}
    \vspace{-7mm}
    \caption{Uniform vector-space B-splines of different orders.
    The control points, which are also exactly the points of the degenerate $k=1$ spline, are shown as diamonds.
    The linear $k = 2$ spline (thin solid line) is the connection between these control points. The quadratic $k = 3$ spline (dashed line) and cubic $k = 4$ spline (solid line) are also shown. Each segment is colored to correspond with the first of the $k$ control points used in its interpolation.}
    \vspace{-3mm}
    \label{fig:spline_diagram}
\end{figure}

\paragraph*{Vector Space}
Through $\mathbf{M}_i$, the \emph{blending matrix}, one can compute the coefficients $\lambda_{i, j}(t)$ of the \emph{blending vector} $\bm{\lambda}_i(t) = \mathbf{M}_i \mathbf{u}_i(t) \in \mathbb{R}^k$, that enable interpolation on the $i$-th segment of the curve.
Interpolation on the $i$-th segment is simply
\begin{equation}
    \mathbf{x}_i(t) = \sum_{j = 0}^{k-1} \lambda_{i, j}(t) \mathbf{x}_{i +j}
    .
    \label{eqn:spline_interpolation:noncumulative}
\end{equation}
\Cref{fig:spline_diagram} illustrates some 2-dimensional vector-space splines.

\paragraph*{Lie Groups}
To efficiently work with Lie group splines, the \emph{cumulative} interpolation formulation must be used~\cite{sommer2020efficient}.
The \emph{cumulative blending matrix} $\widetilde{\mathbf{M}}_i$ is defined as a cumulative sum over columns of $\mathbf{M}_i$,
$\widetilde{m}_{j,n} := \sum_{s = j}^{k-1} m_{s,n}$,
and $\widetilde{\lambda}_{i, j}(t)$ are the coefficients of the \emph{cumulative blending vector} $\bm{\widetilde{\lambda}}_i(t) = \widetilde{\mathbf{M}}_i \mathbf{u}_i(t) \in \mathbb{R}^k$.
Then, using the composition operator $\circ$ (matrix multiplication for many groups), the $\text{Exp}$ map, and the $\text{Log}$ map defined by \cite[Equations~(1), (21) and (22)]{sola2018micro}, the interpolation at $t$ is given as
\begin{gather}
    \mathbf{x}_i(t) = \mathbf{x}_i \circ \mathbf{A}_{i,1}(t) \circ \mathbf{A}_{i,2}(t) \circ \dots \circ \mathbf{A}_{i,k-1}(t) \in \mathcal{L}
    ,
    \label{eqn:spline_interpolation:X}\\
    \text{with}~\mathbf{A}_{i,j}(t) := \text{Exp}(\widetilde{\lambda}_{i,j}(t) \mathbf{d}_j^i) \in \mathcal{L},
    \label{eqn:spline_interpolation:A}\\
    \text{and}~\mathbf{d}_j^i := \mathbf{x}_{i + j} \ominus \mathbf{x}_{i + j - 1} = \text{Log}(\mathbf{x}_{i + j - 1}^{-1} \circ \mathbf{x}_{i + j}) \in \mathbb{R}^d
    .
    \label{eqn:spline_interpolation:d}
\end{gather}

\subsubsection{Spline Types}
\label{sec:theory:temporal_splines:spline_types}

The computation of the as-yet-undefined blending matrix $\mathbf{M}_i$ depends on the spline type, defined by a selection of internal constraints that determine its continuity properties.
The temporal \emph{continuity} $C^n$ of a spline refers to its degree-of-differentiability, where $n$ is the number of derivatives that are continuous over the entire input domain of the spline.
The $i$-th derivative is a piecewise polynomial of degree $m - i$.
Closed-form formulas have been derived for the first three temporal derivatives~\cite{sommer2020efficient}.
$\mathbf{M}_i$ also depends on the knots, which often have \emph{uniform} spacing at some \emph{knot interval}.
Otherwise, the spline is \emph{non-uniform}.

\paragraph*{B-Splines}

The most popular spline used in robotics applications is the \emph{B-spline}, since it has the highest continuity possible; $C^{k-2}$ (not $C^{k-1}$, e.g. in~\cite{cioffi2022continuous,sommer2020efficient,quenzel2021real}).
If the B-spline is uniform, then each element at indices $(s,n)$ of the blending matrix can be set using binomial coefficients~\cite{sommer2020efficient} as 
\begin{equation}
    \begin{gathered}
    m_{s,n} = \frac{\begin{pmatrix}k - 1 \\ n\end{pmatrix}}{(k-1)!} \sum_{l = s}^{k-1} (-1)^{l-s} \begin{pmatrix}k \\ l - s\end{pmatrix} (k-1-l)^{k-1-n}
    .
    \end{gathered}
    \label{eqn:blending_matrix_entries}
\end{equation}
The blending matrix of non-uniform splines depends on $2m$ knots (not $2k$,  e.g., in~\cite{furgale2015continuous}).
It is the transpose of the \emph{basis matrix} $\mathbf{B}_i$ from \citet{haarbach2018survey}, originally \citet{qin1998general},
and so it can be computed by transposing the result of Qin's recursive formula. 
\Cref{tab:spline_types} summarizes common spline types.

\begin{table*}
    \centering
    \caption{Common Spline Types and their Properties.}
    \vspace{-3mm}
    \resizebox{\textwidth}{!}{
    \def\arraystretch{0.9}
    \rowcolors{2}{gray!8}{white}
    \begin{tabular}{|l|ll|l|l|}
        \hline
        \textbf{Spline} & \textbf{Continuity} & \textbf{Interpolating} & \textbf{Example Applications} & \textbf{Additional Notes}\\
        \hline
        B\'ezier & $C^{k-4}$/$C^{k-3}$ & Every $m$-th & Vector Graphics, Fonts & $C^{k-3}$ if control points are mirrored about knot points.\\
        Hermite & $C^{k-4}$/$C^{k-3}$ & All & Animation & Velocity set at control points. Special case of B\'ezier spline. \\
        Kochanek-Bartels & $C^{k-4}$/$C^{k-3}$ & All & Computer Graphics & Modification of Hermite Spline.\\
        Cardinal & $C^{k-3}$ & All & Animation & Special case of Kochanek-Bartels spline.\\
        Catmull-Rom & $C^{k-3}$ & All & Animation, Smoothing & Special case of Cardinal spline.\\
        Linear & $C^0$ & All & Fast Interpolation & Special case of spline with order $k=2$.\\
        B-Spline & $C^{k-2}$ & None in general & Animation, Camera Paths & Minimal support w.r.t. order, smoothness and domain.\\
        \acs{NURBS} & $C^{k-2}$ & None in general & \ac{CAD} & Special case of B-spline.\\
        \hline
    \end{tabular}
    }
    \vspace{-5mm}
    \label{tab:spline_types}
\end{table*}

\subsubsection{Optimization}
\label{sec:theory:temporal_splines:optimization}

In formulating a spline optimization problem, all priors and measurements must be included as \emph{interpolated factors}.
This means that interpolation (\Cref{eqn:spline_interpolation:X}) must be used in the residual computation to obtain the state or its temporal derivatives at the required sample times.
Therefore, sufficient control points and knots are required to perform interpolation at all prior and measurement times.
As a result, these factors in the optimization are associated with \textit{i)} the control points of the spline segment(s) upon which interpolation will be performed, and \textit{ii)} any necessary time-invariant variables.
Jacobians for these factors, required for optimization, can be computed analytically~\cite{sommer2020efficient,tirado2022jacobian,li2023continuous,johnson2024continuous}.

\subsection{\acfp{TGP}}
\label{sec:theory:temporal_gaussian_processes}

In contrast to splines, which model the state parametrically as a weighted combination of control points, \acp{TGP} are non-parametric, using instead a \acf{GP}:
\begin{equation}
    \mathbf{x}(t) \sim \mathcal{GP}(\mathbf{\check{x}}(t), \mathbf{\check{P}}(t, t'))
    .
\end{equation}
Here $\mathbf{\check{x}}(t)$ is the prior mean function and $\mathbf{\check{P}}(t, t')$ is the prior covariance function, or \emph{kernel}.
Interestingly, \acp{TGP} can be considered a weighted combination of infinitely many \acfp{TBF}, made non-parametric by substituting the mean and covariance with functions (known as the \textit{kernel trick})~\cite{tong2012gaussian,tong2013gaussian}.
Indeed, despite different design choices, \ac{CT} methods (based on \acp{TBF}) are sometimes related.
For example, posterior interpolation with a \acs{WNOA} prior (\Cref{sec:theory:temporal_gaussian_processes:gp_priors}) yields cubic Hermite spline interpolation~\cite{barfoot2017state}.
As in \ac{DT} estimation, the estimated process variables $\mathbf{x}$ represent the state at specific times.
Similarly to spline knots, the times of these \emph{support states} may be chosen freely (uniformly or non-uniformly).
\looseness=-1

This method requires the evolution between these support states to be modeled as a \acs{LTV}-\acs{SDE} process as introduced in \Cref{sec:state_estimation_in_robotics}.
While this imposes a Markovian restriction on the \ac{CT} state, augmenting the base state with derivatives to overcome this is usually straightforward, shown later in \Cref{sec:theory:temporal_gaussian_processes:gp_priors}.
Since many systems of interest in robotics can be modeled as \acs{LTV}-\acsp{SDE}, this method is quite general.

It is assumed that the process is governed by \ac{CT} white noise
\begin{equation}
    \mathbf{w}(t) \sim \mathcal{GP}(\mathbf{0}, \mathbf{Q}_C \cdot \delta(t - t'))
    ,
    \label{eqn:gp_pm_noise}
\end{equation}
where $\mathbf{Q}_C$ is the (symmetric, positive-definite) \emph{power-spectral density matrix} and $\delta(\cdot)$ is the \emph{Dirac delta} function.
The mean function $\mathbf{\check{x}}(t) = \mathbb{E}[\mathbf{x}(t)]$ can be derived from \Cref{eqn:ltvsde:solution} as
\begin{gather}
    \mathbf{\check{x}}(t) = \mathbf{\Phi}(t, t') \mathbf{\check{x}}(t') + \mathbf{v}(t, t')
    ,
    \label{eqn:gp_prior_mean_function}\\
    \text{with}~\mathbf{v}(t, t') := \int_{t'}^t \mathbf{\Phi}(t, s) \mathbf{v}(s) ds
    .
\end{gather}
From this it is easy to formulate a residual $\mathbf{e}_i$ between two explicitly estimated states at times $t_i$ and $t_{i+1}$ as
\begin{equation}
    \mathbf{e}_i = \mathbf{\Phi}(t_{i+1}, t_{i}) \mathbf{x}(t_{i}) + \mathbf{v}(t_{i+1}, t_i) - \mathbf{x}(t_{i+1})
    .
    \label{eqn:gp_binary_prior_residual}
\end{equation}
The covariance function $\mathbf{\check{P}}(t, t') = \text{Var}[\mathbf{x}(t)]$ can also be derived from \Cref{eqn:ltvsde:solution} (cf.~\cite{barfoot2014batch,anderson2015batch,barfoot2017state,mukadam2018continuous}) as
\begin{gather}
    \resizebox{0.89\linewidth}{!}{$\displaystyle
    \mathbf{\check{P}}(t, t') = \mathbf{\Phi}(t, t_0) \mathbf{\check{P}}_0 \mathbf{\Phi}(t', t_0)^T + \begin{cases}
            \mathbf{\Phi}(t, t') \mathbf{Q}(t', t_0) & t' < t\\
            \mathbf{Q}(t, t_0) & t' = t\\
            \mathbf{Q}(t, t_0) \mathbf{\Phi}(t', t)^T & t' > t
        \end{cases}
    $}
    \label{eqn:gp_prior_covariance_term}\\
    \text{with}~\mathbf{Q}(t, t') := \int_{t'}^{t} \mathbf{\Phi}(t, s) \mathbf{L}(s) \mathbf{Q}_C \mathbf{L}(s)^T \mathbf{\Phi}(t, s)^T ds
    .
    \label{eqn:gp_Q(t_t')}
\end{gather}
Together, the residual $\mathbf{e}_i$ with corresponding prior covariance $\mathbf{Q}(t_{i+1}, t_i)$ are used as a binary factor between adjacent states, commonly referred to as the \emph{\acs{GP} (motion) prior factor}.

\subsubsection{Interpolation}
\label{sec:theory:temporal_gaussian_processes:interpolation}

\paragraph*{Vector Space}

The key insight is that the process model has been used to define the prior mean and covariance functions of the \ac{GP}.
Because the process model is Markovian, the inverse covariance is \emph{exactly sparse} (block-tridiagonal), and by exploiting this sparsity~\cite{barfoot2014batch,anderson2015batch}, the interpolated posterior mean and covariance can be derived\footnote{
The \ac{GP} interpolation equations can also be derived from the perspective of optimal control theory~\cite[appendix A]{johnson2024continuous}.
} for $t \in [t_i, t_{i+1})$ as
\begin{gather}
    \resizebox{0.88\linewidth}{!}{$\displaystyle
    \mathbf{x}(t) = \mathbf{\Lambda}(t) \mathbf{x}(t_i) + \mathbf{\Psi}(t) \mathbf{x}(t_{i+1}) + \mathbf{v}(t, t_i) - \mathbf{\Psi}(t)\mathbf{v}(t_{i+1}, t_i)
    $}
    ,
    \label{eqn:gp_posterior_mean_simplified}\\
    \resizebox{0.89\linewidth}{!}{$\displaystyle
    \begin{aligned}
    \mathbf{P}(t, t) &= \begin{bmatrix}\mathbf{\Lambda}(t) & \mathbf{\Psi}(t)\end{bmatrix} \begin{bmatrix}\mathbf{P}(t_i, t_i) & \mathbf{P}(t_i, t_{i+1})\\\mathbf{P}(t_{i+1}, t_i) & \mathbf{P}(t_{i+1}, t_{i+1})\end{bmatrix} \begin{bmatrix}\mathbf{\Lambda}(t)^T\\ \mathbf{\Psi}(t)^T\end{bmatrix}\\
    &\quad + \mathbf{Q}(t, t_i) - \mathbf{\Psi}(t) \mathbf{Q}(t_{i+1}, t_i) \mathbf{\Psi}(t)^T
    ,
    \end{aligned}
    $}
    \label{eqn:gp_posterior_covariance_simplified}
\end{gather}
\vspace{-5mm}
\begin{align}
    \text{with}~\mathbf{\Psi}(t) &:= \mathbf{Q}(t, t_i) \mathbf{\Phi}(t_{i + 1}, t)^T \mathbf{Q}(t_{i+1}, t_i)^{-1},
    \label{eqn:tgp_Psi} \\
    \text{and}~\mathbf{\Lambda}(t) &:= \mathbf{\Phi}(t, t_{i}) - \mathbf{\Psi}(t)\mathbf{\Phi}(t_{i + 1}, t_i).
    \label{eqn:tgp_Lambda}
\end{align}
Importantly, interpolation relies only on the two temporally adjacent states.
If $\mathbf{v}(t) = \mathbf{0}, \forall t \in [t_i, t_{i+1})$ as in commonly used models, then \Cref{eqn:gp_posterior_mean_simplified} further simplifies to
\begin{equation}
    \mathbf{x}(t) = \mathbf{\Lambda}(t) \mathbf{x}(t_i) + \mathbf{\Psi}(t) \mathbf{x}(t_{i+1})
    .
    \label{eqn:gp_posterior_mean_simplified_v=0}
\end{equation}

\paragraph*{Lie Groups}

To work with Lie group states, which would yield a nonlinear \ac{SDE} process in almost all non-trivial cases, the method of \emph{local} \ac{LTV}-\acp{SDE} is used~\cite{anderson2015full,dong2017sparse,dong2018sparse}, which utilizes the local tangent space of the manifold.
Let $\mathbf{x}(t) = \{\bm{x}(t), \dot{\bm{x}}(t), \ddot{\bm{x}}(t), \dots\}$ be the global state, where $\bm{x}(t) \in \mathcal{L}$ is a Lie group element (e.g., pose), and $\dot{\bm{x}}(t) \in \mathbb{R}^d, \ddot{\bm{x}}(t) \in \mathbb{R}^d$ are its first and second temporal derivatives. 
Between every pair of explicitly estimated states, $\mathbf{x}_i := \mathbf{x}(t_i)$ and $\mathbf{x}_{i+1} := \mathbf{x}(t_{i+1})$, the \ac{NTV}-\ac{SDE} is approximated as an \ac{LTV}-\ac{SDE} for $t \in [t_i, t_{i+1})$ through the use of a local mapping
\begin{equation}
    \resizebox{0.89\linewidth}{!}{$\displaystyle
    \mathbf{\Xi}(t) := \begin{bmatrix}
        \bm{\xi}(t) \\
        \dot{\bm{\xi}}(t) \\
        \ddot{\bm{\xi}}(t) \\
        \vdots
    \end{bmatrix} := \begin{bmatrix}
        \bm{x}(t) \ominus \bm{x}(t_i) \\
        \mathbf{J}_r(\bm{\xi}(t))^{-1}\dot{\bm{x}}(t) \\
        \frac{d}{dt} \left(\mathbf{J}_r(\bm{\xi}(t))^{-1}\right)\dot{\bm{x}}(t) + \mathbf{J}_r(\bm{\xi}(t))^{-1}\ddot{\bm{x}}(t) \\
        \vdots
    \end{bmatrix}
    $}
    \label{eqn:gp_global_to_local}
\end{equation}
where $\mathbf{J}_r(\bm{\xi}(t))$ is the right Jacobian of the manifold as defined by \cite[Equation~(67)]{sola2018micro}.\footnote{
Note that $\frac{d}{dt} \left(\mathbf{J}_r(\bm{\xi}(t))^{-1}\right)$ does not have a general closed form, but can be approximated~\cite{tang2019white}.
A closed form solution exists for $SO(3)$~\cite{nguyen2024gptr}.
}
The local-to-global mapping is
\begin{gather}
    \bm{x}(t) = \bm{x}(t_i) \oplus \bm{\xi}(t) \in \mathcal{L}
    ,
    \label{eqn:gp_local_to_global_base}\\
    \begin{bmatrix}
        \dot{\bm{x}}(t) \\
        \ddot{\bm{x}}(t) \\
        \vdots
    \end{bmatrix} = \begin{bmatrix}
        \mathbf{J}_r(\bm{\xi}(t)) \dot{\bm{\xi}}(t) \\
        \mathbf{J}_r(\bm{\xi}(t)) \left(\ddot{\bm{\xi}}(t) - \frac{d}{dt} \left(\mathbf{J}_r(\bm{\xi}(t))^{-1}\right) \dot{\bm{x}}(t)\right) \\
        \vdots
    \end{bmatrix}
    .
    \label{eqn:gp_local_to_global_derivatives}
\end{gather}
Therefore, one stores and optimizes the global state variables $\mathbf{x}$ but uses the local \ac{LTV}-\ac{SDE} for interpolation.
Interpolating the state mean at $t \in [t_i, t_{i+1})$ is then a three-step process:
\begin{enumerate}
    \item Map explicitly estimated global states $\mathbf{x}(t_i), \mathbf{x}(t_{i+1})$ to local states $\mathbf{\Xi}(t_i), \mathbf{\Xi}(t_{i+1})$ using \Cref{eqn:gp_global_to_local}.
    \item Interpolate local state $\mathbf{\Xi}(t)$ using \Cref{eqn:gp_posterior_mean_simplified} or \eqref{eqn:gp_posterior_mean_simplified_v=0}.
    \item Map the interpolated local state $\mathbf{\Xi}(t)$ to a global state $\mathbf{x}(t)$ using \Cref{eqn:gp_local_to_global_base,eqn:gp_local_to_global_derivatives}.
\end{enumerate}
A similar procedure is available for the covariance~\cite{anderson2017batch}.

\subsubsection{\acs{GP} Priors}
\label{sec:theory:temporal_gaussian_processes:gp_priors}


The \acf{WNOA} or \textit{`constant-velocity'} prior is a generic process model and was the first to be used for \ac{GP}-based \ac{CT} state estimation~\cite{tong2012gaussian}.
Other priors are discussed in \Cref{sec:survey:temporal_gaussian_processes}.
It assumes the second derivative (`acceleration') is driven by white noise (\Cref{eqn:gp_pm_noise}).
For vector-valued $\bm{x}(t)$, this is written as
\begin{equation}
    \ddot{\bm{x}}(t) = \mathbf{w}(t) \in \mathbb{R}^{d}
    .
\end{equation}
A first-order \ac{SDE} (satisfying the Markovian condition) is formulated by including the first derivative in the state:
\begin{equation}
    \mathbf{x}(t) = \begin{bmatrix}\bm{x}(t)\\\dot{\bm{x}}(t)\end{bmatrix} \in \mathbb{R}^{2d}, ~\text{and}~
    \dot{\mathbf{x}}(t) = \begin{bmatrix}\dot{\bm{x}}(t) \\ \ddot{\bm{x}}(t)\end{bmatrix} \in \mathbb{R}^{2d}
    .
\end{equation}
This yields a \ac{LTI} \ac{SDE} with
\begin{equation}
    \begin{gathered}
    \mathbf{F}(t) = \begin{bmatrix}\mathbf{0} & \mathbf{I}\\\mathbf{0} & \mathbf{0}\end{bmatrix} 
    , ~
    \mathbf{v}(t) = \mathbf{0} 
    , ~\text{and}~
    \mathbf{L}(t) = \begin{bmatrix}\mathbf{0}\\\mathbf{I}\end{bmatrix} 
    .
    \end{gathered}
    \label{eqn:wnoa_FvL}
\end{equation}
The transition, covariance and precision matrices for the \acs{LTI}-\acs{SDE} are calculated in closed form (with $\Delta t := t - t'$) as
\begin{gather}
    \mathbf{\Phi}(t, t') 
    = \begin{bmatrix}\mathbf{I} & \Delta t \mathbf{I}\\
    \mathbf{0} & \mathbf{I}\end{bmatrix} \in \mathbb{R}^{2d \times 2d}
    ,\\
    \mathbf{Q}(t, t') = \begin{bmatrix}
    \frac{1}{3} \Delta t^3 \mathbf{Q}_C & \frac{1}{2} \Delta t^2 \mathbf{Q}_C\\
    \frac{1}{2} \Delta t^2 \mathbf{Q}_C & \Delta t \mathbf{Q}_C 
    \end{bmatrix} \in \mathbb{R}^{2d \times 2d}
    , ~\text{and}~\\
    \mathbf{Q}(t, t')^{-1} = \begin{bmatrix}
    12 \Delta t^{-3} \mathbf{Q}_C^{-1} & -6 \Delta t^{-2} \mathbf{Q}_C^{-1}\\
    -6 \Delta t^{-2} \mathbf{Q}_C^{-1} & 4 \Delta t^{-1} \mathbf{Q}_C^{-1} 
    \end{bmatrix} \in \mathbb{R}^{2d \times 2d}
    .
\end{gather}
Since $\mathbf{v}(t) = \mathbf{0}, \forall t$, the computation of the \ac{GP} prior residual is simplified, and \Cref{eqn:gp_posterior_mean_simplified_v=0} can be used.

If $\bm{x}(t)$ is a Lie group object, then one must have the global Markov state $\mathbf{x}(t) = \{\bm{x}(t), \dot{\bm{x}}(t)\} \in \mathcal{L} \times \mathbb{R}^d$.
The local Markov state defined at $t \in [t_i, t_{i+1})$ and its temporal derivative are
\begin{equation}
    \mathbf{\Xi}(t) = \begin{bmatrix}\bm{\xi}(t)\\\dot{\bm{\xi}}(t)\end{bmatrix} \in \mathbb{R}^{2d}, ~\text{and}~\dot{\mathbf{\Xi}}(t) = \begin{bmatrix}\dot{\bm{\xi}}(t) \\ \ddot{\bm{\xi}}(t)\end{bmatrix} \in \mathbb{R}^{2d}
    .
\end{equation}
The local \ac{WNOA} prior thus assumes that acceleration \emph{in the local frame} is driven by white noise, as
$\ddot{\bm{\xi}}(t) = \mathbf{w}(t) \in \mathbb{R}^d$.
Substituting into \Cref{eqn:gp_binary_prior_residual}, the \ac{WNOA} prior residual (with $\Delta t_{i+1} := t_{i+1} - t_i$) can be written as
\begin{equation}
    \mathbf{e}_i = \begin{bmatrix}
    \Delta t_{i+1} \dot{\bm{x}}(t_i) - \bm{x}_{i+1} \ominus \bm{x}_i \\ \dot{\bm{x}}(t_i) - \mathbf{J}_r(\bm{x}_{i+1} \ominus \bm{x}_i)^{-1} \dot{\bm{x}}(t_{i+1})
    \end{bmatrix}
    .
    \label{eqn:wnoa_prior_residual}
\end{equation}

\vspace{-2mm}



\section{Survey}
\label{sec:survey}



\subsection{Interpolation \& Integration}
\label{sec:survey:interpolation_and_integration}

An overview of the considered works that use interpolation and integration in a \ac{CT} formulation is provided in \Cref{tab:li_ct_applications_robotics}.

\subsubsection{\acf{LI} in \acs{LIDAR} and RADAR Systems}

Among the earliest works that formulate \ac{LI} as a \ac{CT} method are the \acs{LIDAR} \acs{SLAM} system by \citet{bosse2009continuous} and Zebedee~\cite{bosse2012zebedee}, which apply the constant-velocity assumption over short time intervals for pose interpolation in their surfel-based \acs{LIDAR} and \acs{LIDAR}-inertial \ac{SLAM} systems, respectively.
Due to its speed, many works have adopted \ac{LI} for point cloud \acf{MDC} in \ac{LIDAR} odometry~\cite{hong2010vicp,moosmann2011velodyne,dong2013lighting,zhang2014loam,zhang2017low,deschaud2018imls,lin2020loam,vizzo2023kiss}, \ac{LIDAR}-inertial odometry~\cite{ye2019tightly,shan2020lio,li2021towards}, \ac{LIDAR}-inertial \ac{SLAM}~\cite{nguyen2023slict}, \ac{LIDAR}-visual-inertial odometry~\cite{shan2021lvi}, and inertial-wheeled odometry~\cite{droeschel2017continuous}.
\citet{ceriani2015pose} model an $SE(3)$ trajectory with \ac{LI} in their \acs{LIDAR} \ac{SLAM} system, repeatedly correcting for motion distortion in its scan-to-map \acs{ICP}-based optimizer.
FAST-LIO~\cite{xu2021fast} performs a Riemann-style summation of \acs{IMU} measurements in its \ac{LIO} system, both forward, for state propagation in the \ac{KF}, and backward, for point cloud \ac{MDC}.
\ac{LI} has also been applied to \ac{MDC} of spinning RADAR~\cite{vivet2013localization,burnett2021we}.

The \ac{LO} systems CT-ICP~\cite{dellenbach2022ct} and ECTLO~\cite{zheng2023ectlo} embrace the \ac{CT} formulation for \ac{LIDAR} \ac{MDC} within the residuals of iterative optimization, as does the \acs{LIDAR}-inertial \ac{SLAM} system by \citet{park2018elastic} in its surfel and inertial residuals.
Later surfel-based \ac{SLAM} works by the same authors~\cite{park2021elasticity,ramezani2022wildcat} maintain \ac{LI} as the primary representation for fast inference, but use cubic B-splines to obtain pose corrections.
\citet{park2020spatiotemporal} apply \ac{LI} over $SE(3)$ poses to the spatial-temporal calibration of \acs{LIDAR}-camera systems.
Recently, \citet{zheng2024trajlo} formulate point-to-plane residuals for every point of a scanning \ac{LIDAR} in their \ac{LO} system, using \ac{LI} to estimate the $SE(3)$ pose at the acquisition time of each point.

\subsubsection{\acf{LI} in Camera Systems}

Many works have also applied \ac{LI} to camera systems, especially in \acf{RSC}.
An early example was by \citet{klein2009parallel}, who applied \ac{LI} to perform \ac{RSC} within \acs{PTAM}~\cite{klein2007parallel}.
The constant velocity was determined using the feature correspondences, and a similar idea was adopted by the \acs{RANSAC} approach of \citet{anderson2013ransac}.
\ac{RSC} was very common in early \ac{CT} works, through \ac{LI}~\cite{ait2009structure,forssen2010rectifying,ringaby2011scan,jia2012probabilistic,hedborg2012rolling,guo2014efficient} or Riemannian integration of gyroscope measurements~\cite{karpenko2011digital}.
\ac{LI} has also been used to efficiently infer poses for multi-sensor fusion~\cite{geneva2018asynchronous}, as well as in multi-camera odometry and calibration systems~\cite{geneva2018asynchronous,eckenhoff2019multi}.

\subsubsection{Arguments for \acf{LI}}

A common technique present in many \ac{CT} \ac{LI} works~\cite{bosse2009continuous,bosse2012zebedee,zlot2013efficient,zlot2014efficient,dube2016non,park2021elasticity,ramezani2022wildcat} is the separation of a base trajectory of discrete poses and a correction trajectory during estimation.
The key idea is that when using \ac{LI}, corrections can be adequately represented at a lower frequency than the poses themselves~\cite{dube2016non,park2021elasticity}, with few coefficients to be optimized.
Furthermore, \ac{LI} is simple to implement with low computational cost.

\begin{table*}[ht]
    \centering
    \caption{\acf{LI} and numerical integration \ac{CT} state representation works. See \Cref{tab:variable_definitions} for definitions.}
    \vspace{-3mm}
    \resizebox{\textwidth}{!}{
    \def\arraystretch{0.95}
    \rowcolors{2}{gray!8}{white}
    \begin{tabular}{|l|l|l|l|ll|}
        \hline
         \textbf{Authors} & \textbf{Year} & \textbf{Applications} & \textbf{Time Invariant Estimates} & \multicolumn{2}{l|}{\textbf{Time-Varying Estimates}}\\
         &               &                       & State(s) & State(s) & State Representation(s) \\
        \hline
        \citet{ait2009structure} & 2009 & \acs{RS} Object Tracking & $\mathbf{L}$ & $\dot{\mathbf{r}}, \dot{\mathbf{p}}$ & \acl{LI} \\
        \citet{bosse2009continuous} & 2009 & \acs{LIDAR} \acs{SLAM} & $\mathcal{M}_S$ & $\mathbf{T} \in SO(3) \times \mathbb{R}^3$ & \acl{LI} (\& Cubic Spline) \\
        \citet{klein2009parallel} & 2009 & \acs{RS} \acs{SLAM} & $\mathbf{L} $ & $\mathbf{T} \in SE(3)$ & Riemann Summation \\
        \citet{forssen2010rectifying} & 2010 & \acs{RSC} & - & $\mathbf{T} \in SO(3) \times \mathbb{R}^3$ & \acl{LI} \\
        \citet{hong2010vicp} & 2010 & \acs{PC} \acs{MDC} for \acs{PC} Registration & - & $\mathbf{T} \in SE(3)$ & \acl{LI} \\
        \citet{moosmann2011velodyne} & 2011 & \acs{PC} \acs{MDC} for \acs{LO} & - & $\mathbf{T} \in SO(3) \times \mathbb{R}^3$ & \acl{LI} \\
        \citet{ringaby2011scan} & 2011 & \acs{RSC} & - & $\mathbf{T} \in SO(3) \times \mathbb{R}^3$ & \acl{LI} \\
        \citet{karpenko2011digital} & 2011 & \acs{RSC} for Video Stabilization & $\mathbf{K}$ & $\mathbf{r} \in SO(3)$ & \acl{LI} \\
        \citet{hedborg2012rolling} & 2012 & \acs{RSC} for \acs{SFM} & $\mathbf{L}$ & $\mathbf{T} \in SO(3) \times \mathbb{R}^3$ & \acl{LI} \\
        \citet{bosse2012zebedee} & 2012 & \acs{LIDAR}-Inertial \acs{SLAM} & $\mathbf{b}_g, \mathbf{b}_a, \mathbf{t}_E, \mathcal{M}_S$ & $\mathbf{T} \in SO(3) \times \mathbb{R}^3$ & \acl{LI} \\
        \citet{jia2012probabilistic} & 2012 & \acs{RSC} for Video Rectification & - & $\mathbf{r} \in SO(3)$ & \acl{LI} \\
        \citet{anderson2013ransac} & 2013 & \acs{LO} & - & \begin{tabular}{@{}l@{}}$\dot{\mathbf{T}}$\\ $\mathbf{T} \in SE(3)$\end{tabular} & \begin{tabular}{@{}l@{}}Feature Tracking + RANSAC\\ Riemann Summation\end{tabular} \\
        \citet{vivet2013localization} & 2013 & \begin{tabular}{@{}l@{}}RADAR \acs{MDC} for RADAR Odometry \\ Linear Velocity Estimation\end{tabular} & - & ${}^2\dot{\mathbf{T}}$ & \begin{tabular}{@{}l@{}}\acl{LI} \\ \ac{TBF}\end{tabular} \\
        \citet{dong2013lighting} & 2013 & \acs{PC} \acs{MDC} for \acs{PC} Registration for \acs{LO} & - & $\mathbf{T} \in SO(3) \times \mathbb{R}^3$ & \acl{LI} \\
        \citet{li2013real} & 2013 & \acs{VIO} & - & \begin{tabular}{@{}l@{}}$\mathbf{r} \in SO(3)$\\ $\mathbf{p}, \dot{\mathbf{p}}$\\ $\mathbf{b}_g, \mathbf{b}_a$\end{tabular} & \begin{tabular}{@{}l@{}}Euler Method (\acs{RK})\\ Trapezoidal Rule (\acs{RK})\\ Discrete-Time Values\end{tabular} \\
        \citet{guo2014efficient} & 2014 & \acs{RS} Visual-Inertial \acs{SLAM} & $\mathbf{L}$ & \begin{tabular}{@{}l@{}}$\mathbf{T} \in SU(2) \times \mathbb{R}^3, \dot{\mathbf{p}}$ \\ $\mathbf{b}_g, \mathbf{b}_a$\end{tabular} & \begin{tabular}{@{}l@{}}\acl{LI} \\ Discrete-Time Values\end{tabular} \\
        \citet{zhang2014loam,zhang2017low} & 2014/2017 & \acs{PC} \acs{MDC} for \acs{PC} Registration for \acs{LO} & - & $\mathbf{T} \in SO(3) \times \mathbb{R}^3$ & \acl{LI} \\
        \citet{ceriani2015pose} & 2015 & \acs{LIDAR} \acs{SLAM} & $\mathcal{M}$ & $\mathbf{T} \in SE(3)$ & \acl{LI} \\
        \citet{dube2016non} & 2016 & \acs{LIDAR}-Inertial-Wheeled \acs{SLAM} & $\mathcal{M}_S$ & $\mathbf{T} \in SO(3) \times \mathbb{R}^3$ & Linear Splines \\
        \citet{droeschel2017continuous} & 2017 & \acs{PC} \acs{MDC} for \acs{LIDAR} \acs{SLAM} & - & $\mathbf{T} \in SO(3) \times \mathbb{R}^3$ & \acl{LI} \\
        \citet{deschaud2018imls} & 2018 & \acs{PC} \acs{MDC} for \acs{LIDAR} \acs{SLAM} & - & $\mathbf{T} \in SE(3)$ & \acl{LI} \\
        \citet{wang2018continuous} & 2018 & Stereo \acs{VO} & - & $\ddot{\mathbf{r}}, \dot{\mathbf{r}}, \ddot{\mathbf{p}}$ & Constant-Acceleration Integration \\
        \citet{lowe2018complementary} & 2018 & \acs{LIDAR}-Visual-Inertial \acs{SLAM} & $\mathbf{S_3}$ & $\mathbf{T} \in SO(3) \times \mathbb{R}^3$ & Uniform Linear Spline \\
        \citet{park2018elastic} & 2018 & \acs{LIDAR}-Inertial \acs{SLAM} & $\mathbf{b}_g, \mathbf{b}_a, \mathcal{M}_S$ & $\mathbf{T} \in SO(3) \times \mathbb{R}^3$ & \acl{LI} \\
        \cite{geneva2018asynchronous} & 2018 & Multi-Sensor Odometry and Localization & - & $\mathbf{T} \in SO(3) \times \mathbb{R}^3$ & \acl{LI} \\
        \citet{eckenhoff2019multi} & 2019 & Multi-Camera \acs{VIO} and Calibration & $\mathbf{T}_E \in (SU(2) \times \mathbb{R}^3)^n, \mathbf{t}_E, \mathbf{K}^n, \bm{d}_C^n$ & \begin{tabular}{@{}l@{}}$\mathbf{T} \in SU(2) \times \mathbb{R}^3, \dot{\mathbf{p}}$ \\ $\mathbf{b}_g, \mathbf{b}_a$\end{tabular} & \begin{tabular}{@{}l@{}}\acl{LI} \\ Discrete-Time Values\end{tabular} \\
        \citet{ye2019tightly} & 2019 & \acs{PC} \acs{MDC} for \acs{LO} & - & \begin{tabular}{@{}l@{}}$\mathbf{T} \in SU(2) \times \mathbb{R}^3$ \\ $\mathbf{b}_g, \mathbf{b}_a$\end{tabular} & \begin{tabular}{@{}l@{}}\acl{LI} \\ Discrete-Time Values\end{tabular} \\
        \citet{lin2020loam} & 2020 & \acs{PC} \acs{MDC} for \acs{PC} Registration for \acs{LO} & - & $\mathbf{T} \in SO(3) \times \mathbb{R}^3$ & \acl{LI} \\
        \citet{shan2020lio} & 2020 & \acs{PC} \acs{MDC} for \acs{LIDAR}-Inertial \acs{SLAM} & - & $\mathbf{T} \in SO(3) \times \mathbb{R}^3$ & \acl{LI} \\ 
        \citet{park2020spatiotemporal} & 2020 & \acs{LIDAR}-Visual Calibration & $\mathbf{T}_E, \mathbf{t}_E$ & $\mathbf{T} \in SE(3)$ & \acl{LI} \\
        \citet{park2021elasticity} & 2021 & \acs{LIDAR}-Visual-Inertial \acs{SLAM} & $\mathbf{b}_g, \mathbf{b}_a, \mathbf{t}_E, \mathcal{M}_S$ & $\mathbf{T} \in SE(3) \text{ / } SO(3) \times \mathbb{R}^3$ & \acl{LI} (\& Uniform Cubic B-Spline) \\
        \citet{burnett2021we} & 2021 & Spinning RADAR \acs{MDC} & - & ${}^2\dot{\mathbf{T}}$ & \acl{LI} \\
        \citet{li2021towards} & 2021 & \acs{PC} \acs{MDC} for \acs{LIO} & - & \begin{tabular}{@{}l@{}}$\mathbf{p}$ \\ $\mathbf{r} \in SU(2), \mathbf{b}_g, \mathbf{b}_a$\end{tabular} & \begin{tabular}{@{}l@{}}\acl{LI} \\ Discrete-Time Values\end{tabular} \\
        \citet{wang2021online} & 2021 & Visual-Inertial-\acs{LIDAR} Extrinsic Calibration & $\mathbf{T}_E$ & \begin{tabular}{@{}l@{}}$\mathbf{r} \in SU(2), \dot{\mathbf{p}}$ \\ $\mathbf{p}$ \\ $\mathbf{b}_g, \mathbf{b}_a$\end{tabular} & \begin{tabular}{@{}l@{}}\acl{LI} \\ Constant-Acceleration Integration \\ Discrete-Time Values\end{tabular} \\
        \citet{shan2021lvi} & 2021 & \acs{PC} \acs{MDC} for \acs{LIDAR}-Visual-Inertial \acs{SLAM} & - & $\mathbf{T} \in SO(3) \times \mathbb{R}^3$ & \acl{LI} \\ 
        \citet{xu2021fast} & 2021 & \acs{LIO} \& \acs{PC} \acs{MDC} & $\mathbf{g}$ & \begin{tabular}{@{}l@{}}$\mathbf{T} \in SO(3) \times \mathbb{R}^3, \dot{\mathbf{p}}$\\ $\mathbf{b}_g, \mathbf{b}_a$\end{tabular} & \begin{tabular}{@{}l@{}}Forward/Backward Riemann Summation\\ Discrete-Time Values\end{tabular} \\
        \citet{xu2022fast} & 2022 & \acs{LIO} & $\mathbf{g}$ & \begin{tabular}{@{}l@{}}$\mathbf{T} \in SO(3) \times \mathbb{R}^3, \dot{\mathbf{p}}$\\ $\mathbf{b}_g, \mathbf{b}_a$\end{tabular} & \begin{tabular}{@{}l@{}}Constant Angular Velocity \& Acceleration Integration \\ Discrete-Time Values\end{tabular} \\ 
        \citet{wang2022mvil} & 2022 & \acs{PC} \acs{MDC} for \acs{LIDAR}-Visual-Inertial \acs{SLAM} & - & $\mathbf{T} \in SU(2) \times \mathbb{R}^3$ & \acl{LI} \\
        \citet{ramezani2022wildcat} & 2022 & \acs{LIDAR}-Inertial \acs{SLAM} & $\mathbf{b}_g, \mathbf{b}_a, \mathcal{M}_S$ & $\mathbf{T} \in SO(3) \times \mathbb{R}^3$ & \acl{LI} (\& Cubic B-Spline) \\
        \citet{dellenbach2022ct} & 2022 & \acs{PC} \acs{MDC} for \acs{PC} Registration for \acs{LO} & - & $\mathbf{T} \in SO(3) \times \mathbb{R}^3$ & \acl{LI} \\
        \citet{he2023point} & 2023 & \acs{LIO} & $\mathbf{g}$ & \begin{tabular}{@{}l@{}}$\mathbf{T} \in SO(3) \times \mathbb{R}^3, \dot{\mathbf{p}}$\\ $\mathbf{b}_g, \mathbf{b}_a$\end{tabular} & \begin{tabular}{@{}l@{}}Constant Angular Velocity \& Acceleration Integration \\ Discrete-Time Values\end{tabular} \\
        \revision{\citet{vizzo2023kiss}} & \revision{2023} & \acs{PC} \acs{MDC} for \ac{LO} & \revision{-} & \revision{$\mathbf{T} \in SO(3) \times \mathbb{R}^3$} & \revision{Linear Interpolation} \\
        \citet{nguyen2023slict} & 2023 & \acs{PC} \acs{MDC} for \acs{LIDAR}-Inertial \acs{SLAM} & - & $\mathbf{r} \in SO(3), \mathbf{p}, \dot{\mathbf{p}}, \mathbf{b}_g, \mathbf{b}_a$ & \acl{LI} \\
        \citet{zheng2023ectlo} & 2023 & \acs{LO} & - & $\mathbf{T} \in SE(3)$ & \acl{LI} \\
        \citet{chen2023direct} & 2023 & \acs{PC} \acs{MDC} for \acs{LIO} & - & $\mathbf{T} \in SU(2) \times \mathbb{R}^3, \dot{\mathbf{p}}, \mathbf{b}_g, \mathbf{b}_a$ & Constant-Jerk Integration \\
        \citet{zheng2024trajlo} & 2024 & \acs{LO} & - & $\mathbf{T} \in SE(3)$ & \acl{LI} \\
        \hline
    \end{tabular}
    }
    \vspace{-5.5mm}
    \label{tab:li_ct_applications_robotics}
\end{table*}

\subsubsection{Nonlinear Interpolation}

Not all nonlinear interpolation strategies are generalized by the temporal splines or \acp{TGP}.
In the pose interpolation method for \ac{SLAM} proposed by \citet{terzakis2024efficient}, they regress a quadratic rational function to the previous five or more poses with equality constraints to enforce interpolation.
Rational functions take the form of an algebraic fraction where both numerator and denominator are polynomials (i.e., the quotient of quadratic \acp{TBF} in this case).
\revision{\citet{zhu2022chevopt} represent the time-varying state by a \emph{Chebyshev polynomial}, where the temporal base coefficients are regressed through the \emph{collocation method} for batch and sliding-window \ac{MAP} estimation.
They later apply the method to attitude estimation~\cite{zhu2023inertial}.
Given known dynamics, \citet{agrawal2021continuous} use a Chebyshev polynomial basis to represent the \ac{CT} state and control input, leveraging pseudo-spectral control theory for estimation.}

\subsubsection{Numerical Integration}

Integrating the derivative of a state until a time of interest is sometimes interpreted as \ac{CT} state estimation.
This includes Riemannian summation, which follows the same constant rate assumption as \ac{LI}.
In many \acs{IMU} integration~\cite{hol2010modeling,li2013real,xu2021fast} and \acs{IMU} preintegration methods~\cite{forster2015imu,forster2016manifold,shen2015tightly,qin2018vins,brossard2021associating}, analytic equations are derived for the propagation of rotation under a constant angular velocity model, and position under a constant acceleration model.
Occasionally, other models such as constant velocity~\cite{xu2021fast} or constant jerk~\cite{chen2023direct} are assumed for inertial position integration.
FAST-LIO2~\cite{xu2022fast} and Point-LIO~\cite{he2023point} are two notable Kalman filters with continuous kinematic models achieving state-of-the-art performance.
Many other works have employed \acs{IMU} integration (e.g., in \acsp{EKF}~\cite{mirzaei2008kalman,hol2010modeling,bloesch2017iterated} and \acsp{UKF}~\cite{kelly2011visual,kelly2014general}), using a wide variety of numerical integration techniques (e.g., Euler method, trapezoidal rule for \acsp{ODE}~\cite{li2013real}, and \nth{4}-order \acs{RK}~\cite{mirzaei2008kalman,kelly2011visual}).
However, they do not use interpolation to infer intermediate states in a \ac{CT} sense for any purpose.
\citet{wang2018continuous} derive discretized integration equations from assumptions of constant linear and angular acceleration over short time segments.
Treating these as a \ac{CT} representation of a camera pose, they formulate reprojection residuals and optimize for the constant rate parameters of these segments, as well as an ``abrupt force'' parameter to determine when new segments should be created.

\subsubsection{\acl{GP} \acs{IMU} Measurement Preintegration}
\label{sec:survey:interpolation_and_integration:gp_imu_preintegration}

In contrast to the other discussed methods, one collection of works, summarized in \Cref{tab:gp_preintegration_works}, model\revision{s} inertial measurements from an \acs{IMU} with temporal \acp{GP} \revision{to form} preintegrated measurements.
\citet{le20183d} fit \acp{GP} to the measurements and use them to upsample the \ac{IMU} measurements to a very high frequency.
Assuming constant acceleration, \acp{UPM} are computed from these \acp{GP} and a \nth{1}-order method is provided to correct \ac{IMU} biases and inter-sensor time offsets after integration.
They applied \acp{UPM} to \acs{LIDAR}-\acs{IMU} calibration, and later to \acs{LIDAR}-inertial \ac{SLAM}~\cite{le2019in2lama}.
Subsequently~\cite{le2020gaussian}, they presented \acp{GPM}, an analytical method for position, velocity, and single-axis rotation estimation based on linear operators applied to \ac{GP} kernels~\cite{sarkka2011linear} that does not suffer from numerical integration error.
They also derive postintegration biases and time offset corrections for \acp{GPM}.
They applied \acp{UPM} and \acp{GPM} to their \revision{\acs{SLAM}} system IN2LAAMA~\cite{le2020in2laama} and to line-based event-inertial \ac{SLAM}~\cite{le2020idol}.
In follow-up work, \citet{le2021continuous} propose \acp{UGPM} to extend \acp{GPM} to $SO(3)$ rotations, and \acp{LPM} which use \ac{LI} for computational speed.
\citet{dai2022tightly} propose a corner-based event-inertial odometry and mapping system using \acp{LPM} for pose inference at the event times.
Most recently, \citet{le2023continuous} formulate \acp{UGPM} for $SE(3)$ poses, fusing the rotation and acceleration computation steps.
They propose initializing \acp{UGPM} with \acp{LPM}, with an optional kernel hyperparameter tuning step\revision{. They demonstrate their efficacy} in a wide variety of multi-sensor estimation problems \revision{and later use \acp{LPM} for \acs{LIDAR} \ac{MDC} and dynamic object detection~\cite{le2024real}.}
These \ac{GP}-based \ac{IMU} preintegration methods have been shown to be more accurate than other methods~\cite{forster2015imu, forster2016manifold}, especially during fast motion.
\revision{The cubic complexity of \acs{GP} regression in these methods might be overcome with exactly sparse kernels (\Cref{sec:theory:temporal_gaussian_processes}) in future work.}

\begin{table}[t]
    \centering
    \caption{Gaussian process \acs{IMU} measurement preintegration works.}
    \vspace{-3mm}
    \resizebox{\linewidth}{!}{
    \rowcolors{2}{gray!8}{white}
    \begin{tabular}{|l|l|l|}
        \hline
        \textbf{Authors} & \textbf{Year} & \textbf{Contributions} \\
        \hline
        \citet{le20183d} & 2018 & \acsp{UPM} \\
        \citet{le2020gaussian} & 2020 & \acsp{GPM} \\
        \citet{le2020idol} & 2020 & Event-Inertial \acs{SLAM} with \acsp{GPM} \\ 
        \citet{le2021continuous} & 2021 & \begin{tabular}{@{}l@{}}$SO(3)$ \acsp{UGPM} \\ \acsp{LPM}\end{tabular} \\
       \citet{dai2022tightly} & 2022 & Event-Inertial Odometry \& Mapping with \acsp{LPM} \\ 
        \citet{le2023continuous} & 2023 & \begin{tabular}{@{}l@{}}$SE(3)$ \acsp{UGPM} \\ Optional Kernel Hyperparameter Tuning\end{tabular} \\
        \revision{\citet{le2024real}} & \revision{2024} & \revision{\acsp{LPM} for \acs{LIDAR} \acs{MDC} \& Dyn. Obj. Detection} \\
        \hline
    \end{tabular}
    }
    \vspace{-5mm}
    \label{tab:gp_preintegration_works}
\end{table}

\begin{table*}
    \centering
    \caption{Spline and \acf{TBF} \ac{CT} state representation works. Asterisks (*) indicate works that significantly expanded the mathematical formulation of these methods. See \Cref{tab:variable_definitions} for definitions.}
    \vspace{-3mm}
    \resizebox{\textwidth}{!}{
    \def\arraystretch{0.85}
    \rowcolors{2}{gray!8}{white}
    \begin{tabular}{|l|l|l|l|ll|}
        \hline
         \textbf{Authors} & \textbf{Year} & \textbf{Applications} & \textbf{Time-Invariant Estimates} & \multicolumn{2}{l|}{\textbf{Time-Varying Estimates}} \\
        & & & State(s) & State(s) & State Representation(s) \\
        \hline
        \citet{kim1995ac,kim1995general}* & 1995 & Torque Computation, Animation & - & $\mathbf{r} \in SU(2)$ & B\'ezier, Hermite \& B-Spline\\
        \citet{gortler1995hierarchical}* & 1995 & Theory & - & $\mathbb{R}^n$ & Polynomial and Wavelet \acsp{TBF} \\
        \citet{anderson1996spline} & 1996 & Bearing-Based Object Tracking & - & ${}^2\mathbf{p}$ & Cubic B-Spline \\
        \citet{qin1998general}* & 1998 & Theory & - & $\mathbb{R}^n$ & B-Spline\\
        \citet{kang1999cubic} & 1999 & Theory & - & $\mathbf{r}$ & Cubic Splines \\
        \citet{crouch1999casteljau} & 1999 & Theory & - & $\mathcal{L}$ & Cubic Splines \\
        \citet{jung2001camera} & 2001 & \acs{VIO} & \begin{tabular}{@{}l@{}}-\\ -\end{tabular} & \begin{tabular}{@{}l@{}}$\mathbf{r} \in SO(3)$\\ $\mathbf{p}$\end{tabular} & \begin{tabular}{@{}l@{}}Riemann Summation \\ Spline\end{tabular}\\
        \citet{bibby2010hybrid} & 2010 & Dynamic \acs{SLAM} & ${}^2\mathcal{M}_O$ & ${}^2\mathbf{T} \in \mathbb{R}^3, \revision{({}^2\mathbf{T}_D \in \mathbb{R}^3)^D}$ & Non-Uniform Cubic Splines \\
        \citet{hadzagic2011bayesian} & 2011 & Object Tracking & - & ${}^2\mathbf{p}$ & Non-Uniform Cubic B-Spline \\
        \citet{fleps2011optimization} & 2011 & Visual-Inertial Calibration & $\mathbf{b}_g, \mathbf{b}_a, \mathbf{T}_E, \mathbf{g}, \dot{\mathbf{p}}_\text{scale}$ & $\mathbf{p} \in \mathbb{R}^3, \mathbf{r} \in \mathbb{R}^4$ & Cubic B-Splines \\
        \citet{furgale2012continuous,furgale2015continuous}* & 2012/2015 & \begin{tabular}{@{}l@{}}Theory\\ Visual-Inertial Calibration\\ \acs{RS} Localization\end{tabular} & \begin{tabular}{@{}l@{}}-\\$\mathbf{T}_E, \mathbf{g}, \mathbf{L}$ \\-\end{tabular} & \begin{tabular}{@{}l@{}}$\mathbb{R}^n$\\ $\mathbf{r} \in \mathbb{R}^3 \text{(\acs{CGR})}, \mathbf{p}, \mathbf{b}_g, \mathbf{b}_a$\\$\mathbf{r} \in \mathbb{R}^3 \text{(\acs{CGR})}, \mathbf{p}$\end{tabular} & \begin{tabular}{@{}l@{}}\acsp{TBF}\\ B-Splines\\ B-Splines\end{tabular} \\
        \citet{furgale2013unified} & 2013 & Visual-Inertial Calibration & $\mathbf{T}_E, \mathbf{t}_E, \mathbf{g}$ & \begin{tabular}{@{}l@{}}$\mathbf{T} \in \mathbb{R}^6 \text{(\acs{CGR})}$\\ $\mathbf{b}_g, \mathbf{b}_a$\end{tabular} & \begin{tabular}{@{}l@{}}\acsp{TBF} / Quintic B-Spline\\\acsp{TBF} / Cubic B-Splines\end{tabular} \\
        \citet{lovegrove2013spline}* & 2013 & \begin{tabular}{@{}l@{}}\acs{RS} \acs{SLAM}\\\acs{RS}-Inertial \acs{SLAM}\end{tabular} & \begin{tabular}{@{}l@{}}$\bm{\rho}_d$\\ $\mathbf{b}_g, \mathbf{b}_a, \mathbf{K}, \mathbf{T}_E, \mathbf{g}, \bm{\rho}_d$\end{tabular} & $\mathbf{T} \in SE(3)$ & Cubic B-Spline \\
        Anderson et al.~\cite{anderson2013towards,anderson2015relative} & 2013/2015 & \acs{LIDAR} \acs{SLAM} & $\mathbf{L}$ & \begin{tabular}{@{}l@{}}$\dot{\mathbf{T}}$\\$\mathbf{T} \in SE(3)$\end{tabular} & \begin{tabular}{@{}l@{}}\acsp{TBF} / Cubic B-Spline\\Riemann Summation\end{tabular} \\
        \citet{oth2013rolling} & 2013 & \acs{RS} Calibration & $t_\text{RS}$ & $\mathbf{r} \in \mathbb{R}^3, \mathbf{p} \in \mathbb{R}^3$ & \acsp{TBF} / Uniform Cubic B-Spline \\
        \citet{sheehan2013continuous} & 2013 & \acs{LIDAR} Localization & - & $\mathbf{T} \in \mathbb{R}^6 \text{(\acs{RPY})}$ & Cubic Catmull-Rom Spline \\
        \citet{zlot2013efficient,zlot2014efficient} & 2013/2014 & \acs{LIDAR}-Inertial Odometry & $\mathcal{M}_S$ & $\mathbf{T} \in SO(3) \times \mathbb{R}^3$ & Uniform Linear / Cubic B-Spline \\
        \citet{rehder2014spatio} & 2014 & \acs{LIDAR}-Visual-Inertial Calibration & $\mathbf{T}_E, \mathbf{T}_E, \mathbf{g}$ & $\mathbf{T} \in \mathbb{R}^6 \text{(\acs{CGR})}, \mathbf{b}_g, \mathbf{b}_a$ & \acsp{TBF} / Quintic B-Splines \\
        \citet{alismail2014continuous} & 2014 & \acs{PC} \acs{MDC} for \acs{PC} Registration for \acs{LO} & - & $\mathbf{T} \in \mathbb{R}^6 \text{(\acs{CGR})}$ & Uniform Linear to Quintic B-Spline \\
        \citet{anderson2014hierarchical} & 2014 & \acs{TE} & - & $\mathbf{T} \in \mathbb{R}^6$ & Polynomial and Wavelet \acsp{TBF} \\
        \citet{patron2015spline} & 2015 & \begin{tabular}{@{}l@{}}\acs{RS} \acs{SLAM}\\\acs{RS}-Inertial \acs{SLAM}\end{tabular} & \begin{tabular}{@{}l@{}}$\bm{\rho}_d$\\ $\mathbf{b}_g, \mathbf{b}_a, \mathbf{K}, \mathbf{T}_E, \mathbf{g}, \bm{\rho}_d$\end{tabular} & $\mathbf{T} \in SE(3)$ & Cubic B-Spline \\
        \citet{kerl2015dense} & 2015 & \acs{RS} \acs{RGBD} Odometry & - & $\mathbf{T} \in SE(3)$ & Uniform Cubic B-Spline \\
        \citet{mueggler2015continuous} & 2015 & Event Camera Localization & - & $\mathbf{T} \in SE(3)$ & Uniform Cubic B-Spline \\
        \citet{rehder2016general} & 2016 & \begin{tabular}{@{}l@{}}Extrinsic Calibration Theory \vspace{2mm} \\ \acs{LIDAR}-Visual-Inertial Calibration \vspace{2mm} \end{tabular} & \begin{tabular}{@{}l@{}}$\mathbf{t}_E$ \vspace{2mm} \\ $b_r, \mathbf{T}_E, \mathbf{t}_E, \mathbf{g}, \mathbf{L}_\pi$ \vspace{2mm}\end{tabular} & \begin{tabular}{@{}l@{}} - \\ \begin{tabular}{@{}l@{}}$\mathbf{T} \in \mathbb{R}^6 \text{(\acs{AA})}$\\ $\mathbf{b}_g, \mathbf{b}_a$\end{tabular}\end{tabular} & \begin{tabular}{@{}l@{}}\acsp{TBF} \\ \begin{tabular}{@{}l@{}}Quintic B-Splines\\ Cubic B-Splines\end{tabular}\end{tabular}\\
        \citet{rehder2016extending} & 2016 & Multi-\acs{IMU} Visual-Inertial Calibration & \begin{tabular}{@{}l@{}}$\mathbf{S}_g, \mathbf{M}_g, \mathbf{G}_g, \mathbf{S}_a,$ \\ $\quad \mathbf{M}_a, \mathbf{T}_E, \mathbf{t}_E, \mathbf{g}$\end{tabular} & \begin{tabular}{@{}l@{}}$(\mathbf{T} \in \mathbb{R}^6 \text{(\acs{AA})})^N$\\ $(\mathbf{b}_g\mathbf{b}_a)^N$\end{tabular} & \begin{tabular}{@{}l@{}}Quintic B-Splines\\ Cubic B-Splines\end{tabular} \\
        \citet{kim2016direct} & 2016 & \acs{RS} \acs{SLAM} & $\mathbf{p}_d$ & $\mathbf{T} \in SE(3)$ & Cubic B-Spline \\
        \citet{sommer2016continuous}* & 2016 & \begin{tabular}{@{}l@{}}Theory\\ Attitude Estimation\end{tabular} & - & \begin{tabular}{@{}l@{}}$\mathcal{L}$\\ $\mathbf{r} \in SU(2)$\end{tabular} & B-Spline \\
        \citet{li2016fitting,li2017track,li2018joint} & 2016-2018 & Bearing-based Tracking & - & ${}^2\mathbf{p}$ & Polynomial and Sinusoidal \acsp{TBF} \\
        \citet{rehder2017camera} & 2017 & Visual-Inertial Calibration & \begin{tabular}{@{}l@{}}$t_e, \bm{r}_C, \mathbf{S}_g, \mathbf{M}_g, \mathbf{G}_g, \mathbf{S}_a, \mathbf{M}_a,$ \\ $\quad \mathbf{p}_{ia}, \mathbf{r}_{ga}, \mathbf{T}_E, \mathbf{t}_E, \mathbf{g}, \mathbf{m}_o$\end{tabular} & \begin{tabular}{@{}l@{}}$\mathbf{T} \in \mathbb{R}^6 \text{(AA)}$\\ $\mathbf{b}_g, \mathbf{b}_a$\end{tabular} & \begin{tabular}{@{}l@{}}Uniform Quintic B-Spline\\ Uniform Cubic B-Spline\end{tabular} \\ 
        \citet{vandeportaele2017pose} & 2017 & \acs{RS} Localization & - & $\mathbf{T} \in SE(3)$ & Uniform Cubic B-Spline \\
        \citet{droeschel2018efficient} & 2018 & \acs{PC} \acs{MDC}, \acs{LIDAR} \acs{SLAM} & $\mathcal{M}_S$ & $\mathbf{T} \in SE(3)$ & Cubic B-Spline \\
        \citet{mueggler2018continuous} & 2018 & Event-Inertial Localization & $\mathbf{b}_g, \mathbf{b}_a, \mathbf{r}_\mathbf{g} \in \mathbb{R}^2$ & $\mathbf{T} \in SE(3)$ & Uniform Cubic B-Spline \\
        \citet{ovren2018spline} & 2018 & \acs{RS} Visual-Inertial \acs{SLAM} & $\bm{\rho}_d$ & $\mathbf{T} \in SU(2) \times \mathbb{R}^3$ & Cubic B-Splines \\
        \citet{ovren2019trajectory} & 2019 & \acs{RS} Visual-Inertial \acs{SLAM} & $\bm{\rho}_d$ & $\mathbf{T} \in SU(2) \times \mathbb{R}^3, SE(3)$ & Cubic B-Spline(s) \\
        \citet{sommer2020efficient}* & 2020 & \begin{tabular}{@{}l@{}}Theory\\ \acs{TE}\\ Visual-Inertial Calibration\end{tabular} & \begin{tabular}{@{}l@{}}-\\ -\\$\mathbf{b}_g, \mathbf{b}_a, \mathbf{T}_E, \mathbf{g}$\end{tabular} & \begin{tabular}{@{}l@{}}$\mathcal{L}$\\ $\mathbf{T} \in SO(3) \times \mathbb{R}^3, SE(3)$\\ $\mathbf{T} \in SO(3) \times \mathbb{R}^3, SE(3)$\end{tabular} & \begin{tabular}{@{}l@{}}B-Spline\\Cubic to Quintic B-Spline(s)\\Quartic Uniform B-Spline(s)\end{tabular} \\
        \citet{pacholska2020relax} & 2020 & Certifiable Range-Based Localization & - & ${}^n\mathbf{p}$ & Polynomial and Sinusoidal \acsp{TBF}\\
        \citet{lv2020targetless} & 2020 & \acs{LIDAR}-Inertial Calibration & $\mathbf{b}_g, \mathbf{b}_a, \mathbf{T}_E$ & $\mathbf{T} \in SO(3) \times \mathbb{R}^3$ & Uniform Cubic B-Spline \\
        \citet{hug2020hyperslam} & 2020 & Visual-Inertial \acs{SLAM} & $\mathbf{b}_g, \mathbf{b}_a, \mathbf{g}, \mathbf{L}$ & $\mathbf{T} \in SU(2) \times \mathbb{R}^3$ & Cubic B-Spline(s) \\
        \citet{lv2021clins} & 2021 & \begin{tabular}{@{}l@{}}\acs{PC} \acs{MDC}\\ \acs{LIO}\end{tabular} & \begin{tabular}{@{}l@{}} -\\ $\mathbf{b}_g, \mathbf{b}_a$\end{tabular} & \begin{tabular}{@{}l@{}}$\mathbf{T} \in SO(3) \times \mathbb{R}^3, \dot{\mathbf{p}} \text{ \& } \mathbf{T} \in SO(3) \times \mathbb{R}^3$\\ $\mathbf{T} \in SO(3) \times \mathbb{R}^3$\end{tabular} & \begin{tabular}{@{}l@{}}Riemann Summation \& Uniform B-Splines\\ Uniform B-Splines\end{tabular} \\
        \citet{ng2021continuous} & 2021 & Multi-RADAR Inertial Odometry & - & \begin{tabular}{@{}l@{}}$\mathbf{T}\in SE(3)$ \\ $\mathbf{b}_g, \mathbf{b}_a$\end{tabular} & \begin{tabular}{@{}l@{}}Uniform Cubic B-Spline \\ \acl{DT} Values\end{tabular} \\
        \citet{yang2021asynchronous} & 2021 & Multi-Camera \acs{SLAM} & $\mathbf{L}$ & $\mathbf{T} \in SE(3)$ & Non-Uniform Cubic B-Spline \\
        \citet{quenzel2021real} & 2021 & Surfel Registration for \acs{LO} & - & $\mathbf{T} \in SO(3) \times \mathbb{R}^3$ & Non-Uniform Quadratic B-Splines \\
        \citet{huang2021b} & 2021 & \acs{VO} & $\mathbf{L}$ & \begin{tabular}{@{}l@{}}$\mathbf{T} \in SU(2) \times \mathbb{R}^3$ \\ $\mathbf{T} \in \mathbb{R}^4$ (position \& roll)\end{tabular} & Cubic B-Spline \\
        \citet{persson2021practical}* & 2021 & \begin{tabular}{@{}l@{}}\acs{VO} \\ Visual-Inertial Calibration\end{tabular} & \begin{tabular}{@{}l@{}}- \\ $\mathbf{T}_E$\end{tabular} & $\mathbf{T} \in SU(2) \times \mathbb{R}^3$ & Cubic to Quintic B-Spline \\
        \citet{cioffi2022continuous} & 2022 & Visual-Inertial-\acs{GNSS} \acs{SLAM} & $\mathbf{T}_E, \mathbf{t}_E, \mathbf{g}, \mathbf{L}$ & $\mathbf{T} \in SO(3) \times \mathbb{R}^3, \mathbf{b}_g, \mathbf{b}_a$ & Cubic to Sextic Uniform B-Splines \\
        \citet{hug2022continuous} & 2022 & Stereo Visual-Inertial \acs{SLAM} & $\mathbf{g}, \mathbf{L}$ & $\mathbf{T} \in SU(2) \times \mathbb{R}^3, \mathbf{b}_g, \mathbf{b}_a$ & Uniform Cubic B-Splines \\
        \citet{tirado2022jacobian}* & 2022 & Object Tracking & - & $\revision{(\mathbf{T}_D \in SE(3))^D}$ & Non-Uniform Cubic B-Splines \\
        \citet{mo2022continuous} & 2022 & Visual-Inertial \acs{SLAM} & $\mathbf{K}, \mathbf{r}_\mathbf{g} \in \mathbb{R}^2, s, \bm{\rho}_d$ & \begin{tabular}{@{}l@{}}$\mathbf{T} \in SO(3) \times \mathbb{R}^3$\\ $\bm{r}_C \in \mathbb{R}^2, \mathbf{b}_g, \mathbf{b}_a$\end{tabular} & \begin{tabular}{@{}l@{}}Cubic Splines\\ Discrete-Time Values\end{tabular} \\
        \citet{huai2022continuous} & 2022 & \acs{RS} Visual-Inertial Calibration & \begin{tabular}{@{}l@{}}$t_{\text{\acs{RS}}}, \{\mathbf{S}_g, \mathbf{M}_g, \mathbf{r}_{ga}\} \in \mathbb{R}^{3 \times 3},$ \\ $\quad \mathbf{G}_g, \mathbf{S}_a, \mathbf{M}_a, \mathbf{T}_E, \mathbf{t}_E, \mathbf{r}_\mathbf{g}$\end{tabular} & $\mathbf{T} \in \mathbb{R}^6 \text{(AA)}, \mathbf{b}_g, \mathbf{b}_a$ & Uniform Quintic B-Splines \\
        \citet{wang2022visual} & 2022 & Event-Based \acs{VO} & - & ${}^2\mathbf{p}$ & B-Spline \\
        \citet{lang2022ctrl} & 2022 & Visual-Inertial \acs{SLAM} & $t_{\text{\acs{RS}}}, \bm{\rho}_d$ & \begin{tabular}{@{}l@{}}$\mathbf{T} \in SO(3) \times \mathbb{R}^3$\\ $\mathbf{b}_g, \mathbf{b}_a$\end{tabular} & \begin{tabular}{@{}l@{}}Uniform Cubic B-Splines\\ Discrete-Time Values\end{tabular} \\
        \citet{lv2022observability} & 2022 & \acs{LIDAR}-Inertial Calibration & \begin{tabular}{@{}l@{}}$b_r, s_r, \mathbf{m}_r \in \mathbb{R}^4, \mathbf{S}_g, \mathbf{M}_g,$ \\ $\quad \mathbf{G}_g, \mathbf{S}_a, \mathbf{M}_a, \mathbf{T}_E, \mathbf{t_E}, \mathbf{g}$\end{tabular} & \begin{tabular}{@{}l@{}}$\mathbf{T} \in SO(3) \times \mathbb{R}^3$\\ $\mathbf{b}_g, \mathbf{b}_a$\end{tabular} & \begin{tabular}{@{}l@{}}Uniform Cubic B-Splines\\ Discrete-Time Values\end{tabular} \\
        \citet{lv2023continuous} & 2023 & \acs{LIDAR}-Visual-Inertial \acs{SLAM} & $\mathbf{t}_E, \bm{\rho}_d$ & \begin{tabular}{@{}l@{}}$\mathbf{T} \in SO(3) \times \mathbb{R}^3$\\ $\mathbf{b}_g, \mathbf{b}_a$\end{tabular} & \begin{tabular}{@{}l@{}}Uniform Cubic B-Splines\\ Discrete-Time Values\end{tabular} \\
        \citet{lang2023coco,lang2024gaussian} & 2023\revision{/2024} & \acs{LVIO} & - & \begin{tabular}{@{}l@{}}$\mathbf{T} \in SO(3) \times \mathbb{R}^3$\\ $\mathbf{b}_g, \mathbf{b}_a$\end{tabular} & \begin{tabular}{@{}l@{}}Non-Uniform Cubic B-Splines\\ Discrete-Time Values\end{tabular} \\
        \citet{li2023continuous}* & 2023 & Theory, \acs{UWB}-Inertial Localization & $\mathbf{T}_E, \mathbf{r}_\mathbf{g}$ & $\mathbf{T} \in SU(2) \times \mathbb{R}^3, \mathbf{b}_g, \mathbf{b}_a$ & Uniform Cubic B-Splines \\
        \citet{li2023embedding} & 2023 & \acs{UWB}-Inertial Localization & - & $\mathbf{p}$ & Uniform Cubic B-Spline \\
        \citet{jung2023asynchronous} & 2023 & \acs{MDC} for Multi-\acs{LIDAR} Inertial Odometry & - & $\mathbf{T} \in SE(3)$ & Uniform Cubic B-Spline \\
        \revision{\citet{lu2023event}} & \revision{2023} & \revision{Event-Inertial Odometry} & \revision{-} & \revision{\begin{tabular}{@{}l@{}}$\mathbf{p}$ \\ $\mathbf{b}_g, \mathbf{b}_a$\end{tabular}} & \revision{\begin{tabular}{@{}l@{}}Uniform Cubic B-Splines \\ Discrete-Time Values\end{tabular}} \\
        \citet{nguyen2024eigen} & 2024 & \acs{LIO} & - & \begin{tabular}{@{}l@{}} $\mathbf{r} \in SO(3), \mathbf{p}$ \\ $\mathbf{b}_g, \mathbf{b}_a$ \end{tabular} & \begin{tabular}{@{}l@{}} Uniform Cubic B-Splines \\ Discrete-Time Values \end{tabular} \\
        \citet{lv2024cta} & 2024 & \acs{LO} & - & $\mathbf{r} \in SO(3), \mathbf{p}$ & Uniform Cubic B-Splines \\
        \citet{hug2024hyperion} & 2024 & \begin{tabular}{@{}l@{}}Odometry \\ Localization\end{tabular} & - & $\mathbf{T} \in SU(2) \times \mathbb{R}^3$ & Uniform B-/Z-Splines \\
        \citet{li2024continuous} & 2024 & \acs{LIDAR}-RADAR-inertial Odometry and Calibration & $\mathbf{T}_E, \mathbf{t}_E$ & \begin{tabular}{@{}l@{}}$\mathbf{T} \in SU(2) \times \mathbb{R}^3$ \\ $\mathbf{b}_g, \mathbf{b}_a$\end{tabular} & \begin{tabular}{@{}l@{}}Uniform B-Splines \\ Discrete-Time Values\end{tabular} \\
        \revision{\citet{nguyen2024mcd}} & \revision{2024} & \revision{Ground Truth Trajectory Estimation} & \revision{$\mathbf{b}_g, \mathbf{b}_a$} & \revision{$\mathbf{T} \in SO(3) \times \mathbb{R}^3$} & \revision{Uniform Quintic B-Spline} \\
        \revision{\citet{li2024hcto}} & \revision{2024} & \revision{\acs{LIO}} & \revision{-} & \revision{\begin{tabular}{@{}l@{}}$\mathbf{T} \in SU(2) \times \mathbb{R}^3$ \\ $\mathbf{b}_g, \mathbf{b}_a$\end{tabular}} & \revision{\begin{tabular}{@{}l@{}}Uniform Cubic B-Splines \\ Discrete-Time Values\end{tabular}} \\
        \hline
    \end{tabular}
    }
    \vspace{-5.5mm}
    \label{tab:spline_ct_applications_robotics}
\end{table*}

\subsection{Temporal Splines}
\label{sec:survey:temporal_splines}

A chronologically ordered overview of all surveyed methods using temporal splines is given in \Cref{tab:spline_ct_applications_robotics}.

\subsubsection{Origins \& Early Formulations}

Splines have received significant attention in robotics as a \ac{CT} representation of the state.
They were first formulated as piecewise-polynomial by Schoenberg in 1946~\cite{schoenberg1988contributions}, inspired by `lofting' where thin wooden strips (called splines) were used for aircraft and ship design, including for construction from templates during World War II~\cite{bartels1995introduction}.
While various spline types have been explored in different robotics applications, B-splines, short for \textit{basis} splines, are most popular due to their smoothness properties.
\revision{Their mathematical foundations were} pioneered by Schoenberg, de Boor, and others, with the de Boor-Cox recurrence formula~\cite{de1972calculating,de1978practical} a famous early formulation.
From this, \citet{qin1998general} was the first to derive a general matrix representation for B-splines.
Later, splines were presented as a special case of \acp{TBF} where the bases are polynomials~\cite{furgale2015continuous,furgale2012continuous}.
As an alternative to polynomials, wavelet basis functions~\cite{anderson2014hierarchical,gortler1995hierarchical} have been proposed in a hierarchical scheme to account for the varying richness of motion.
Sinusoidal temporal bases have also been used in conjunction with polynomials for certifiable range-based localization~\cite{pacholska2020relax} \revision{and} object tracking~\cite{li2016fitting,li2017track,li2018joint}.

\subsubsection{Generalization to Lie Groups}

Importantly for robotics, splines were formulated for orientations~\cite{kim1995ac,kim1995general,kang1999cubic} and then general Lie groups~\cite{crouch1999casteljau}.
Despite this, many early applications retained a vector spline representation for orientation or pose~\cite{fleps2011optimization,furgale2012continuous,furgale2015continuous,furgale2013unified,oth2013rolling,sheehan2013continuous,rehder2014spatio,alismail2014continuous,rehder2016general,rehder2016extending} before $SE(3)$ splines~\cite{lovegrove2013spline,patron2015spline,mueggler2015continuous,kim2016direct} and general Lie group splines of arbitrary order~\cite{sommer2016continuous} were applied.
Transitioning away from the $\mathbb{R}^3$ formulation for orientation was necessary because of the presence of singularities, the loss of $C^{k-2}$ continuity~\cite{kim1995ac,kim1995general,lovegrove2013spline}, and because interpolation does not represent a minimum distance in rotation space~\cite{lovegrove2013spline,vandeportaele2017pose}.
For these reasons, the resultant curves under the vector formulation may not realistically approximate motion.
Manifold constraints were rarely enforced before the Lie group formulation, although \citet{fleps2011optimization} did so in their \acs{SQP} approach.

\subsubsection{Choice of Pose Representation}

For 3D pose estimation using B-splines, a choice exists between the \emph{joint} ($SE(3)$) and \emph{split} ($SO(3) \times \mathbb{R}^3$ or $SU(2) \times \mathbb{R}^3$) representations.
\citet{ovren2019trajectory} scrutinize this choice for robotic applications and advocate for the split representation in most cases. 
They showed that under the $SE(3)$ representation, translation, and orientation are coupled in \emph{screw motion}, where implicitly ``the pose is manipulated by an internal force and torque'', an action called a \emph{wrench}~\cite{murray2017mathematical}, which may be appropriate in some contexts such as \revision{robot arm end effector} trajectories.
In contrast, they argue that the split representation implicitly assumes that an external force and torque actuate the pose.
Furthermore, they demonstrate that the analytic acceleration of the joint $SE(3)$ spline is not the kinematic body acceleration in general.
Besides the theoretical justifications, they experimentally show that the split representation converges faster and more accurately for their \ac{RS} Visual-Inertial \ac{SLAM} system.
\citet{tirado2022jacobian} make a similar argument; however, they advocate for the joint representation as the general case for rigid-body motion.
\citet{hug2020hyperslam} prefer the split representation, emphasizing the increased computational effort required in the $SE(3)$ representation.
For orientation, \citet{haarbach2018survey} and \citet{li2023continuous} advocate for quaternions ($SU(2)$) over rotation matrices ($SO(3)$).
This can be attributed to their space efficiency, requiring four variables instead of nine, and their relative computational efficiency.
For the same reasons, \citet{haarbach2018survey} advocate for the unit dual quaternion $\mathbb{DH}_1$ over $SE(3)$ for B-spline pose interpolation.

\subsubsection{Modern Advancements}

\citet{sommer2020efficient} devise a efficient method for computing Lie group B-spline derivatives, requiring $\mathcal{O}(k)$ instead of $\mathcal{O}(k^2)$ matrix operations for a spline of order $k$.
They further derive a recursive strategy to compute analytic Jacobians for the $SO(3)$ B-spline and its first and second derivatives in $\mathcal{O}(k)$ time.
This \revision{is} extended to other Lie groups~\cite{tirado2022jacobian,li2023continuous} \revision{with} significant computational savings \revision{demonstrated} over automatic or numerical differentiation~\cite{tirado2022jacobian}.
Finally, \citet[Appendix E]{johnson2024continuous} provide analytic Jacobians of general Lie group splines up to the second derivative.
\citet{hug2020hyperslam} contribute analytic Jacobians for the knots, suggesting the possibility that they could be jointly optimized in future work.
Recently, \citet{hug2024hyperion} use SymForce~\cite{martiros2022symforce} to obtain analytic B-spline implementations that are much faster than those derived by \citet{sommer2020efficient}.
\citet{persson2021practical} provide a number of novel theoretical contributions: \textit{i)} constant-velocity extrapolation through new basis functions, \textit{ii)} a necessary condition on knot spacing for quaternion splines to express a maximum angular velocity, \textit{iii)} a simple method to derive and compute integrals of spline derivative functions, \textit{iv)} a theoretical and experimental investigation of different spline regularizers, and \textit{v)} an approach for how these integral-type costs should be sampled.
Spline regularization has been effective in reducing oscillations~\cite{furgale2015continuous,furgale2013unified} and enforcing non-holonomic constraints~\cite{huang2021b}.
Additionally, \citet{cioffi2022continuous} demonstrated with batch optimization that B-splines are more accurate than \ac{DT} for trajectory estimation, especially when estimating for time offsets between sensors.

\subsubsection{Early Applications in Robotics}

Among the first robotic applications of splines were position estimation in object tracking~\cite{anderson1996spline} and \ac{VIO}~\cite{jung2001camera}.
In another early work, \citet{bibby2010hybrid} represented the vehicle's trajectories and dynamic objects in the environment using cubic interpolating splines in a 2D dynamic \ac{SLAM} application.
\citet{hadzagic2011bayesian} used cubic B-splines for 2D position estimation in ship tracking.
Bosse and Zlot~\cite{bosse2009continuous,zlot2014efficient} used cubic B-splines for trajectory fitting and optimization while keeping \ac{LI} as their primary \ac{CT} representation.
Later, \acp{TBF} were applied to the estimation of robot velocity in a feature-based \acs{LIDAR} \ac{SLAM} context~\cite{anderson2013towards,anderson2015relative}, using cubic B-splines in the implementation and a Riemann summation of velocity to obtain the pose estimates.

\subsubsection{Explicit \acf{MDC}}

As in \ac{LI}, B-splines were also used to correct for \ac{MDC}, including in point cloud registration for \ac{LO}~\cite{alismail2014continuous}, in a \acs{LIDAR} \acs{SLAM} system~\cite{droeschel2018efficient} (extending \citet{droeschel2017continuous}), and \ac{RSC} for \acs{RGBD} odometry~\cite{kerl2015dense}.
The \ac{LO} and mapping system presented in \citet{quenzel2021real} uses a temporal B-spline and \acp{GMM} for local-to-map surfel registration.
\citet{lv2021clins} solve the \acs{LIDAR} \ac{MDC} problem for their \ac{LIO} system by first estimating pose and velocity with \ac{IMU} integration before fitting a uniform B-spline and using it to correct the motion-distorted point cloud.
They use the corrected scans in a feature-based \ac{LIO} optimization where a B-spline again represents the trajectory.
\citet{jung2023asynchronous} use cubic B-spline interpolation for motion and temporal compensation of multiple \acsp{LIDAR} in their \ac{LIO} \ac{KF}.

\subsubsection{Localization, Odometry \& \acs{SLAM}}

\citet{sheehan2013continuous} use a Catmull-Rom spline for \acs{LIDAR} localization, and \citet{li2023continuous} use B-splines for \acs{UWB}-inertial localization.
B-splines have been used to process high-rate event camera measurements, which are particularly hard to aggregate for \ac{DT} systems.
This includes event~\cite{mueggler2015continuous} and event-inertial localization~\cite{mueggler2018continuous}, event \acs{VO}~\cite{wang2022visual}\revision{, and map-free event-inertial ego-velocity estimation~\cite{lu2023event}}.
Leveraging prior work~\cite{nguyen2023slict}, \citet{nguyen2024eigen} propose an iterative, parallelized, on-manifold linear solver for real-time surfel-based \ac{LIO}.
\citet{lv2024cta} use a `spot uncertainty model' to weight point-to-plane residuals in their \ac{LO} system.
\citet{ng2021continuous} apply B-splines to RADAR-inertial odometry, using radial velocity measurements from multiple radars to form ego-velocity residuals.
Adopting uniform cubic B-splines for the trajectory, \citet{li2024continuous} jointly optimize \acs{LIDAR} point-to-plane, RADAR Doppler velocity, and inertial residuals for multi-sensor odometry.
B-splines have also been used as the \ac{CT} \revision{trajectory} formulation in \ac{RS} visual~\cite{lovegrove2013spline,patron2015spline,kim2016direct}, \ac{RS} visual-inertial~\cite{lang2022ctrl}, visual-inertial~\cite{ovren2019trajectory,hug2020hyperslam,hug2022continuous}, multi-camera~\cite{yang2021asynchronous}, \acs{LIDAR}~\cite{droeschel2018efficient}, and \acs{LIDAR}-visual-inertial~\cite{lv2023continuous,lang2023coco,lang2024gaussian} \ac{SLAM}.

\subsubsection{Sensor Calibration}

\acp{TBF} and B-splines have been used for sensor calibration, including estimation of \revision{\ac{RS} camera line delay~\cite{oth2013rolling,huai2022continuous,lang2022ctrl}}, visual-inertial~\cite{furgale2013unified} and \acs{LIDAR}-visual-inertial~\revision{\cite{rehder2014spatio,lv2023continuous}} extrinsic calibration, and comprehensive simultaneous intrinsic and extrinsic calibration for \acs{LIDAR}-visual-inertial~\cite{rehder2016general}, multi-\acs{IMU} visual-inertial~\cite{rehder2016extending}, and visual-inertial~\cite{rehder2017camera,persson2021practical} configurations.
\revision{Uniform} quintic B-splines have been applied to \ac{RS} visual-inertial calibration~\cite{huai2022continuous}, and uniform cubic B-splines to targetless \acs{LIDAR}-inertial calibration in structured environments~\cite{lv2020targetless,lv2022observability}.

\subsubsection{Online Optimization}

While many perform batch optimization, some works operate online using \emph{sliding-window} optimization (a.k.a. \emph{fixed-lag smoothing}).
Early works use linear splines (equivalent to \ac{LI})~\cite{zlot2013efficient,zlot2014efficient,dube2016non,lowe2018complementary}, but \citet{zlot2014efficient} also did with cubic B-splines.
\citet{persson2021practical} were among the first to use windowed optimization for real-time odometry.
\revision{\citet{lang2022ctrl} develop B-spline marginalization strategies for \ac{RS} visual-inertial fixed-lag smoothing, refined by \citet{lv2023continuous} to achieve real-time sliding-window \acs{LIDAR}-visual-inertial \ac{SLAM}.}
More recent spline-based odometry systems have continued to develop marginalization strategies~\cite{lang2023coco,lv2024cta,li2024continuous}, with \citet{lang2023coco} to first to do so \revision{for} non-uniform splines.
Importantly, the use of per-point residuals in these works allows \ac{MDC} to occur simultaneously to state estimation~\cite{lv2023continuous,lang2023coco,lv2024cta,li2024continuous}.
\citet{quenzel2021real}, \citet{mo2022continuous} who incorporate inertial measurements through cubic splines,
and \citet{tirado2022jacobian} who employ non-uniform cubic B-splines for object tracking, also employ sliding-window optimization.
In contrast to these optimization approaches, \citet{li2023embedding} devises a probabilistic recursive filter that maintains the control points for a single cubic B-spline segment in its state and can thus run in real-time with uncertainty estimation.
\citet{hug2024hyperion} take a different approach, proposing a \ac{GBP} optimization framework called Hyperion, where factor graph optimization is achieved through efficient message passing between nodes and factors.
While usually slower than existing \ac{NLLS} solvers such as Ceres~\cite{agarwal_ceres_2023}, \ac{GBP} computes variable (inverse) uncertainty during optimization instead of as a post-optimization procedure, allowing strategies for targeted updating (e.g., of non-converged nodes), and is inherently scalable to multi-agent systems.

\begin{table*}
    \centering
    \caption{\acf{TGP} \ac{CT} state representation works. Stars ($^\filledstar$) indicate works that significantly expanded the mathematical formulation of this method. See \Cref{tab:variable_definitions} for definitions.}
    \vspace{-3mm}
    \resizebox{\textwidth}{!}{
    \rowcolors{2}{gray!8}{white}
    \begin{tabular}{|l|l|l|l|ll|}
        \hline
         & \textbf{Year} & \textbf{Applications} & \textbf{Time-Invariant Estimates}    & \multicolumn{2}{l|}{\textbf{Time-Varying Estimates}} \\
         &               &                       & State(s) & State(s) & State Representation(s) \\
        \hline
        \citet{tong2012gaussian}$^\filledstar$ & 2012 & \begin{tabular}{@{}l@{}}Theory\\ Range-Based Localization\end{tabular} & \begin{tabular}{@{}l@{}}-\\-\end{tabular} & \begin{tabular}{@{}l@{}}$\mathbb{R}^n$\\ $\{{}^2\mathbf{T}, {}^2\dot{\mathbf{T}}\} \in \mathbb{R}^6$\end{tabular} & \acs{WNOA} \acs{TGP} \\
        \citet{tong2013gaussian}$^\filledstar$ & 2013 & \begin{tabular}{@{}l@{}}Theory\\ Range-Based \acs{SLAM}\end{tabular} & \begin{tabular}{@{}l@{}}-\\ ${}^2\mathbf{L}$\end{tabular} & \begin{tabular}{@{}l@{}}$\mathbb{R}^n$\\ ${}^2\mathbf{T} \in \mathbb{R}^3$\end{tabular} & \acs{WNOA} \acs{TGP} \\
        \citet{tong2014pose} & 2014 & Laser-Based Visual \acs{SLAM} & $\mathbf{L}$ & $\mathbf{T} \in SE(3)$ & \acs{WNOA} \acs{TGP} \\
        \citet{barfoot2014batch} & 2014 & Range-Based \acs{SLAM} & ${}^2\mathbf{L}$ & $\{{}^2\mathbf{T}, {}^2\dot{\mathbf{T}}\} \in \mathbb{R}^6$ & \acs{WNOA} \acs{TGP} \\
        \citet{yan2014incremental,yan2017incremental}$^\filledstar$ & 2014/2017 & Range-Based \acs{SLAM} & ${}^2\mathbf{L}$ & $\{{}^2\mathbf{T}, {}^2\dot{\mathbf{T}}\} \in \mathbb{R}^6$ & \acs{WNOA} \acs{TGP} \\
        \citet{anderson2015batch}$^\filledstar$ & 2015 & Range-Based \acs{SLAM} & ${}^2\mathbf{L}$ & $\{{}^2\mathbf{T}, {}^2\dot{\mathbf{T}}\} \in \mathbb{R}^6$ & \acs{WNOA} \acs{TGP} \\
        \citet{anderson2015full}$^\filledstar$ & 2015 & Visual \acs{SLAM} & $\mathbf{L}$ & $\mathbf{T} \in SE(3), \dot{\mathbf{T}}$ & \acs{WNOA} \acs{TGP} \\
        \citet{dong20174d} & 2017 & Visual-Inertial-\acs{GNSS} \acs{SLAM} & $\mathbf{L}$ & \begin{tabular}{@{}l@{}}$\mathbf{T} \in SE(3), \dot{\mathbf{T}}$\\ $\mathbf{b}_g, \mathbf{b}_a$\end{tabular} & \begin{tabular}{@{}l@{}}\acs{WNOA} \acs{TGP}\\ Discrete-Time Values\end{tabular} \\
        \citet{dong2017sparse,dong2018sparse}$^\filledstar$ & 2017/2018 & \begin{tabular}{@{}l@{}}Theory\\ Range-Based \acs{SLAM}\\ Inertial Attitude Estimation\\ Monocular Visual \acs{SLAM}\\ Mobile Manipulator Planning (PR2)\end{tabular} & \begin{tabular}{@{}l@{}}-\\ ${}^2\mathbf{L}$\\ -\\ $\mathbf{L}$\\ -\end{tabular} & \begin{tabular}{@{}l@{}}$\mathcal{L}, \mathbb{R}^d$\\ ${}^2\mathbf{T} \in SE(2), \mathbb{R}^3, {}^2\dot{\mathbf{T}} \in \mathbb{R}^3$\\ $\mathbf{r} \in SO(3), \dot{\mathbf{r}}$\\ $\mathbf{T} \in SIM(3), \dot{\mathbf{T}} \in \mathbb{R}^7$\\ $\mathbf{x} \in SE(2) \times \mathbb{R}^{15}, \dot{\mathbf{x}} \in \mathbb{R}^{18}$\end{tabular} & \acs{WNOA} \acs{TGP} \\
        \citet{mukadam2017simultaneous,mukadam2019steap}$^\filledstar$ & 2017/2019 & \begin{tabular}{@{}l@{}}Two-Link Mobile Arm Planning\\ Mobile Manipulator Planning (PR2)\\ Mobile Manipulator Planning (Vector)\end{tabular} & - & \begin{tabular}{@{}l@{}}$\mathbf{x} \in SE(2) \times \mathbb{R}^2, \dot{\mathbf{x}} \in \mathbb{R}^2$\\ $\mathbf{x} \in SE(2) \times \mathbb{R}^{15}, \dot{\mathbf{x}} \in \mathbb{R}^{18}$\\ $\mathbf{x} \in SE(2) \times \mathbb{R}^6, \dot{\mathbf{x}} \in \mathbb{R}^9$\end{tabular} & \acs{WNOA} \acs{TGP} \\
        \citet{rana2018towards} & 2018 & Learning from Demonstration and Motion Planning & - & $\mathbf{p}, \dot{\mathbf{p}}$ & \acs{WNOA}/Learnt \acsp{TGP} \\
        \citet{maric2018singularity} & 2018 & Motion Planning & - & $\mathbf{x} \in \mathbb{R}^n$ & \acs{WNOA} \acs{TGP} \\
        \citet{warren2018towards} & 2018 & \acs{VO} for Teach \& Repeat & - & $\mathbf{T} \in SE(3), \dot{\mathbf{T}}$ & \acs{WNOA} \acs{TGP} \\
        \citet{warren2018there} & 2018 & \acs{VO} for Teach \& Repeat & - & $\mathbf{T} \in SE(3), \dot{\mathbf{T}}$ & \acs{WNOA} \acs{TGP} \\
        \citet{tang2019white}$^\filledstar$ & 2019 & Odometry & - & $\mathbf{T} \in SE(3), \dot{\mathbf{T}}, \ddot{\mathbf{T}}$ & \acs{WNOJ} \acs{TGP} \\
        \citet{judd2019unifying,judd2020occlusion} & 2019/2020 & Dynamic \acs{SLAM} & $\mathbf{L}$ & $\mathbf{T} \in SE(3), \dot{\mathbf{T}}, \revision{(\mathbf{T}_D \in SE(3), \dot{\mathbf{T}}_D)^D}$ & \acs{WNOA} \acsp{TGP} \\
        \citet{barfoot2020exactly} & 2020 & Wheeled Robot \acs{SLAM} & ${}^2\mathbf{L}$ & ${}^2\mathbf{p}, {}^2\mathbf{r} \in \mathbb{R}$ & \acs{WNOA} \acs{TGP} \\
        \citet{wong2020variational} & 2020 & Hyperparameter Estimation for \acs{LIDAR} Localization & - & $\mathbf{T} \in SE(3), \dot{\mathbf{T}}$ & \acs{WNOA} \acs{TGP} \\
        \citet{wong2020data}$^\filledstar$ & 2020 & \begin{tabular}{@{}l@{}}\acs{LIDAR} Localization \\ \acs{LO}\end{tabular} & - & $\mathbf{T} \in SE(3), \dot{\mathbf{T}}, \ddot{\mathbf{T}}$ & Singer \acs{TGP} \\
        \revision{\citet{pervsic2021spatiotemporal}} & \revision{2021} & \revision{Spatiotemporal Calibration from Object Tracking} & \revision{-} & \revision{$(\mathbf{p}_D, \dot{\mathbf{p}}_D, \ddot{\mathbf{p}}_D)^D$} & \revision{\acs{WNOJ} \acsp{TGP}} \\
        \citet{kapushev2021random} & 2021 & \acs{SLAM} & ${}^2\mathbf{L}$ & ${}^2\mathbf{T} \in \mathbb{R}^3$ & \acs{RFF}-Gaussian \acs{TGP}\\
        \citet{judd2021multimotion,judd2024multimotion} & 2021/2024 & Dynamic \acs{SLAM} & $\mathbf{L}$ & $\mathbf{T} \in SE(3), \dot{\mathbf{T}}, \ddot{\mathbf{T}}, \revision{(\mathbf{T}_D \in SE(3), \dot{\mathbf{T}}_D, \ddot{\mathbf{T}}_D)^D}$ & \acs{WNOA} / \acs{WNOJ} \acsp{TGP} \\
        \citet{li20213d} & 2021 & \acs{LIDAR}-Inertial Extrinsic Calibration and Localization & $\mathbf{T}_E, \mathbf{t}_E$ & \begin{tabular}{@{}l@{}}$\mathbf{T} \in SE(3), \dot{\mathbf{T}}$ \\ $\mathbf{b}_g, \mathbf{b}_a$\end{tabular} & \begin{tabular}{@{}l@{}}\acs{WNOA} \acs{TGP}\\ Discrete-Time Values\end{tabular} \\
        \citet{zhang2022onlinefgo} & 2022 & \acs{LIDAR}-Inertial-\acs{GNSS} Odometry & - & \begin{tabular}{@{}l@{}}$\mathbf{T} \in SE(3), \dot{\mathbf{T}}, \ddot{\mathbf{T}}$\\ $\mathbf{b}_g, \mathbf{b}_a, b_{\text{\acs{GNSS}, t}}, b_{\text{\acs{GNSS}, v}}, n_{\text{\acs{GNSS}}}$\end{tabular} & \acs{WNOJ} \acs{TGP} \\
        \citet{zhang2022continuous} & 2022 & Inertial-\acs{GNSS} Odometry & - & \begin{tabular}{@{}l@{}}$\mathbf{T} \in SE(3), \dot{\mathbf{T}}$\\ $\mathbf{b}_g, \mathbf{b}_a, b_{\text{\acs{GNSS}, t}}, b_{\text{\acs{GNSS}, v}}$\end{tabular} & \begin{tabular}{@{}l@{}}\acs{WNOA} \acs{TGP}\\ Discrete-Time Values\end{tabular} \\
        \citet{wu2022picking} & 2022 & \acs{FMCW} \acs{LIDAR} Odometry & - & $\mathbf{T} \in SE(3), \dot{\mathbf{T}}$ & \acs{WNOA} \acs{TGP} \\
        \citet{burnett2022we} & 2022 & \acs{LIDAR} and RADAR Odometry for Teach \& Repeat & - & $\mathbf{T} \in SE(3), \dot{\mathbf{T}}$ & \acs{WNOA} \acs{TGP} \\
        \citet{liu2022asynchronous} & 2022 & Event-Based \acs{SLAM} for \acs{VO} & $\mathbf{L}$ & $\mathbf{T} \in SE(3), \dot{\mathbf{T}}$ & \acs{WNOA} \acs{TGP} \\
        \citet{wang2023event} & 2023 & Event-Based \acs{SLAM} for \acs{VO} & $\mathbf{L}$ & $\mathbf{T} \in SE(3), \dot{\mathbf{T}}$ & \acs{WNOA} \acs{TGP} \\
        \citet{yoon2023need} & 2023 & \acs{FMCW} \acs{LIDAR}-Inertial Odometry & - & $\dot{\mathbf{T}}$ & \acs{WNOA} \acs{TGP} \\
        \revision{\citet{goudar2023continuous}} & \revision{2023} & \revision{\acs{UWB} Localization} & \revision{-} & \revision{$\mathbf{T} \in SE(3), \dot{\mathbf{T}}$} & \revision{\acs{WNOA} \acs{TGP}} \\
        \citet{le2023continuous_2} & 2023 & Event \acs{MDC} Feature Tracking & - & ${}^2\mathbf{T} \in \mathbb{R}^3$ & Squared Exponential \acp{TGP} \\
        \citet{le2023gaussian} & 2023 & IMU-Pose Extrinsic Calibration & $\mathbf{T}_E, \mathbf{t}_E$ & \begin{tabular}{@{}l@{}}$\mathbf{T} \in SO(3) \times \mathbb{R}^3$ \\ $\mathbf{b}_g, \mathbf{b}_a, \mathbf{r}_\mathbf{g} \in SO(3)$ \end{tabular} & \begin{tabular}{@{}l@{}}Gaussian \acs{TGP} \\ Discrete-Time Values\end{tabular} \\
        \citet{zheng2024trajlio} & 2024 & \acs{LIO} & - & \begin{tabular}{@{}l@{}}$\mathbf{r} \in SO(3), \dot{\mathbf{r}}$ \\ $\mathbf{p}, \dot{\mathbf{p}}, \ddot{\mathbf{p}}$ \\ $\{\mathbf{b}_g\}^{N_g}, \{\mathbf{b}_a\}^{N_a}$\end{tabular} & \begin{tabular}{@{}l@{}}\acs{WNOA} \acs{TGP} \\ \acs{WNOJ} \acs{TGP} \\ \acs{WNOV} \acsp{TGP} \end{tabular} \\
        \citet{johnson2024continuous}$^\filledstar$ & 2024 & Visual-Inertial Localization & $\mathbf{b}_g, \mathbf{b}_a$ & \begin{tabular}{@{}l@{}}$\mathbf{T} \in SE(3) / SO(3) \times \mathbb{R}^3, \dot{\mathbf{T}}, \ddot{\mathbf{T}}$ \\ $\mathbf{T} \in SE(3) / SO(3) \times \mathbb{R}^3$\end{tabular} & \begin{tabular}{@{}l@{}} \acs{WNOJ} \ac{TGP}(s) \\ Uniform Cubic / Quintic B-Spline(s)\end{tabular} \\
        \citet{lisus2024doppler} & 2024 & RADAR Odometry for Teach \& Repeat & - & $\mathbf{T} \in SE(3), \dot{\mathbf{T}}$ & \acs{WNOA} \acs{TGP} \\
        \citet{burnett2024continuous} & 2024 & \acs{LIDAR}-Inertial and  RADAR-Inertial Odometry & - & \begin{tabular}{@{}l@{}}$\mathbf{T} \in SE(3), \dot{\mathbf{T}}$ \\ $\mathbf{b}_g, \mathbf{b}_a, \left(\mathbf{r}_{\mathbf{g}} \in SO(3)\right)$\end{tabular} & \begin{tabular}{@{}l@{}}\acs{WNOA} \acs{TGP} \\ \acs{WNOV} \acsp{TGP}\end{tabular} \\
        \citet{barfoot2024certifiably} & 2024 & Certifiable Trajectory Estimation & - & $\mathbf{T} \in SE(3), \dot{\mathbf{T}}$ & \acs{WNOA} \acs{TGP} \\
        \citet{zhang2024gnss} & 2024 & \acs{LIDAR}-Inertial-\acs{GNSS}-Velocimeter Odometry & - & \begin{tabular}{@{}l@{}}$\mathbf{T} \in SE(3), \dot{\mathbf{T}}, \ddot{\mathbf{T}}$\\ $\mathbf{b}_g, \mathbf{b}_a, b_{\text{\acs{GNSS}, t}}, b_{\text{\acs{GNSS}, v}}$\end{tabular} & \begin{tabular}{@{}l@{}}\acs{WNOJ} \acs{TGP}\\ Discrete-Time Values\end{tabular} \\
        \citet{nguyen2024gptr} & 2024 & \begin{tabular}{@{}l@{}}Visual-Inertial Localization and Calibration \\ \acs{UWB}-Inertial Localization \\ Multi-\acs{LIDAR} Localization and Calibration\end{tabular} & $\mathbf{T}_E, \mathbf{g}, \mathbf{b}_g, \mathbf{b}_a$ & $\mathbf{T} \in SO(3) \times \mathbb{R}^3, \dot{\mathbf{T}}, \ddot{\mathbf{T}}$ & \acs{WNOJ} \acsp{TGP} \\
        \revision{\citet{burnett2024imu}} & \revision{2024} & \revision{\acs{LIO}} & \revision{-} & \revision{$\mathbf{T} \in SE(3), \dot{\mathbf{T}}, \ddot{\mathbf{T}}, \mathbf{b}_g, \mathbf{b}_a$} & \revision{Singer \& \acs{WNOV} \acsp{TGP}} \\
        \revision{\citet{shen2024ctemlo}} & \revision{2024} & \revision{Multi-\acs{LIDAR} Odometry} & \revision{-} & \revision{$\mathbf{T} \in SO(3)\times\mathbb{R}^3, \dot{\mathbf{T}}, \ddot{\mathbf{p}}$} & \revision{Constant Angular Velocity \& Acceleration \acs{TGP}} \\
        \hline
    \end{tabular}
    }
    \vspace{-5mm}
    \label{tab:gp_ct_applications_robotics}
\end{table*}

\subsection{\acfp{TGP}}
\label{sec:survey:temporal_gaussian_processes}

A chronologically ordered overview of all the \ac{TGP} methods surveyed in this work is presented in \Cref{tab:gp_ct_applications_robotics}.

\subsubsection{Early Works}

The use of \acp{GP} for \ac{CT} state estimation was first realized by Tong et al.~\cite{tong2012gaussian,tong2013gaussian,tong2013laser}, where they presented \ac{GPGN}, a nonlinear batch optimization method, and applied it to 2D range-based localization~\cite{tong2012gaussian} and \ac{SLAM}~\cite{tong2013gaussian}.
In these early formulations, the connection with \acp{TBF} was made explicit in the derivation, with the transition from parametric to non-parametric achieved through the \emph{kernel trick}.
\citet{tong2014pose} subsequently extended \ac{GPGN} to 3D for their laser-based visual \ac{SLAM} system.
Crucially, \citet{barfoot2014batch} and \citet{anderson2015batch} demonstrated that \acp{GP} using the \acs{LTV}-\acs{SDE} class of prior have exactly sparse inverse kernel matrices, enabling efficient regression and interpolation.
The computation of the inverse kernel matrix is reduced from cubic to linear time, and interpolation can be computed in constant time.
When using this prior with a linear measurement model, \revision{they show exact equivalence} to discrete-time smoothing \revision{with identical computational complexity} when evaluated at the measurement times~\cite{anderson2015batch,barfoot2017state}.
Solving the \ac{SLAM} problem using this method was called \ac{STEAM}, with \citet{anderson2015full} extending the approach to $SE(3)$ poses using local \acs{LTV}-\acsp{SDE} (\Cref{sec:theory:temporal_gaussian_processes:interpolation}).
Nonlinear models are supported in these works through linearization and iterative optimization.

\subsubsection{Optimization}

Exploiting the sparsity of the formulation, the \ac{STEAM} problem can be modeled as a factor graph with binary \ac{GP} prior factors connecting adjacent states~\cite{anderson2015batch,anderson2015full}\revision{, enabling batch~\cite{anderson2015full} and sliding-window optimization~\cite{goudar2023continuous}}.
\citet{yan2014incremental,yan2017incremental} used this to develop a \ac{GP} version of the incremental optimizer \acs{ISAM2}~\cite{kaess2012isam2}, and with \acs{IGPMP2}~\cite{mukadam2018continuous} applied this incremental \ac{GP} approach to motion-planning problems.
\emph{Sparse interpolated measurement factors} allow measurements to be added between the explicitly estimated states~\cite{tong2013gaussian,anderson2015batch,yan2017incremental,anderson2017batch}, and can be readily used within an incremental Bayes tree optimization scheme~\cite{yan2017incremental}.
\citet{dong2017sparse, dong2018sparse} provided a formulation for general matrix Lie groups before integrating it into their \ac{STEAP} framework~\cite{mukadam2017simultaneous, mukadam2019steap}.
Exploring a different direction, \citet{barfoot2020exactly} propose \ac{ESGVI} to fit a Gaussian to the full Bayesian posterior of the problem states. 
This approximation ought to be more representative than the Laplace approximation from \ac{MAP} estimation (\Cref{eqn:laplace_approximation}).
They devise a Newton-style iterative solver that avoids the need for analytic derivatives through cubature rules and exploit\revision{s} the inherent sparsity of the problem structure to achieve the same computational complexity as \ac{MAP} estimation.
\citet{wong2020variational} then extend this approach for hyperparameter learning with an \ac{EM} optimization framework.

\subsubsection{Applications using the \acs{WNOA} Prior}

The \ac{GN} \ac{TGP} approach~\cite{anderson2015batch,anderson2015full,dong2017sparse} has been applied in the context of \textit{teach and repeat}, such as in \ac{VO} for \acsp{UAV}~\cite{warren2018towards,warren2018there}, or \ac{LIDAR} and RADAR odometry~\cite{burnett2022we,lisus2024doppler}.
\citet{liu2022asynchronous} and \citet{wang2023event} have proposed incremental event-based \ac{VO} pipelines based on this \ac{CT} representation, employing the \ac{GN} algorithm in the underlying optimization.
\citet{li20213d} demonstrate \ac{LIDAR}-inertial extrinsic calibration and localization in structured environments.
Incremental \revision{Bayes-tree-based} optimizers~\cite{kaess2012isam2} have been used in \ac{SLAM} for agricultural monitoring~\cite{dong20174d} and inertial-\acs{GNSS} odometry~\cite{zhang2022continuous}.
Recent works~\cite{wu2022picking,yoon2023need} have applied \acp{TGP} to \acs{FMCW} \ac{LIDAR} \revision{odometry}.
Rather than aggregating points \revision{(e.g., for scan registration)}, \citet{wu2022picking} formulate point-to-plane and radial-velocity residuals for each point.
\citet{yoon2023need} combine radial-velocity residuals with a gyroscope for linear estimation of the total \acs{DOF} ego-velocity, which they integrate into $SE(3)$ poses.
Judd and Gammell~\cite{judd2020occlusion,judd2019unifying} address the \emph{multimotion} estimation problem with \acp{TGP} in their occlusion-aware \ac{VO} pipeline, estimating for the $SE(3)$ ego-motion and trajectories of points on dynamic objects.

\subsubsection{Applications using Other Priors}

The aforementioned works all use the \ac{WNOA} prior.
\citet{tang2019white} propose the \emph{`constant-acceleration'} \ac{WNOJ} model instead.
They achieve higher accuracies in some cases and prove that when the chosen prior does not match the motion, the state estimate will be biased along certain \acs{DOF}.
The \ac{WNOJ} prior has since been used for online \acs{GNSS} multi-sensor odometry~\cite{zhang2022onlinefgo,zhang2024gnss} \revision{and target-based multi-sensor extrinsic calibration~\cite{pervsic2021spatiotemporal}}.
Recently, \citet{zheng2024trajlio} propose multi-sensor \ac{LIO} using \ac{WNOJ} for position, \ac{WNOA} for orientation, and \ac{WNOV} (a.k.a. `random-walk') for \ac{IMU} biases, and achieve state-of-the-art results.
As in \citet{wu2022picking} and their previous work~\cite{zheng2024trajlo}, they avoid measurement aggregation (e.g., \ac{LIDAR} registration, \ac{IMU} pre-integration) while retaining real-time performance through the split pose representation and efficient map management.
\citet{burnett2024continuous} also utilize the \ac{WNOV} prior for \ac{IMU} biases, demonstrating effective \ac{LIDAR}-inertial and RADAR-inertial odometry on various datasets.
\citet{judd2021multimotion,judd2024multimotion} compare the \ac{WNOJ} and \ac{WNOA} priors in their \ac{MVO} pipeline~\cite{judd2019unifying,judd2020occlusion}.
\citet{johnson2024continuous} recently were the first to compare \acp{TGP} and B-splines in visual-inertial localization experiments.
They compared \ac{WNOJ} \acp{TGP} to cubic and quintic B-splines, both with and without motion prior factors, and for both the split and joint pose representations.
They conclude that \revision{with} the same motion models and measurements\revision{,} and \revision{when} the degree-of-differentiability matches, temporal splines and \acp{GP} are similar in accuracy and solve time.
Furthermore, they show that motion prior factors act as regularizers, preventing overfitting to noisy measurements.
\revision{However, these benefits are significantly diminished when an \ac{IMU} is available.}
\citet{zheng2024trajlo} and \citet{johnson2024continuous} investigate the efficacy of the split pose representation and conclude its primary benefit is the reduction of computation time\revision{, both at solve and inference time}.
\revision{From the existing literature, \citet{johnson2024continuous} provide insights into computation timings of these \ac{CT} method variants, while \citet{cioffi2022continuous} demonstrate that \ac{CT} solve times can be on par with \ac{DT}.}
Most recently, \citet{nguyen2024gptr} \revision{released} their split pose \acs{WNOJ} implementation and provide localization and calibration examples for a variety of sensor combinations.
They further provide a generalization of \ac{WNOD} priors for any order and analytic Jacobians for their residuals.
\citet{wong2020data} investigate the \emph{Singer prior}, which represents latent accelerations as a \ac{GP} with a Mat\'ern kernel.
They argue that the data-driven optimization of an additional parameter enables a better representation of the robot's motion.
\revision{\citet{burnett2024imu} utilize the Singer prior for \ac{LIO}, and comparatively analyze the use of \ac{IMU} as a motion model input (for preintegration) and as a measurement.}

\subsubsection{Other Approaches}

\revision{\citet{shen2024ctemlo} adapt the \ac{TGP} method to an \acs{EKF} for the first time to achieve real-time dropout-resilient multi-\ac{LIDAR} odometry. They create point-to-voxel residuals with a degeneracy-aware point sampling strategy and an incremental, probabilistic voxel-based map.}
\citet{le2023continuous_2} propose a novel method for event camera motion compensation using \acp{GP}.
By treating camera motion as $SE(2)$ motion in the image frame over short time periods, they model each \acs{DOF} as independent zero-mean \acp{GP} with Gaussian kernels.
The inducing values (i.e., set of observations) of these are then the hyperparameters of another \ac{GP} representing a continuous occupancy field of events in the image plane.
By optimizing these through log-marginal likelihood maximization, this occupancy field, and hence a continuous distance field as its negative logarithm, can be inferred and used for homography registration and pattern tracking.
\citet{le2023gaussian} use a \ac{GP} pose representation in the extrinsic calibration of an \ac{IMU} with \revision{a} pose measurement reference frame (e.g., motion capture or robotic arm frames).
\citet{kapushev2021random} propose a \ac{SLAM} algorithm evaluated in 2D which models the pose as a \ac{GP}\revision{,} with a \ac{RFF} approximation of the Gaussian kernel and a discretized motion model or smoothing splines for the prior mean.

\section{Open Problems}
\label{sec:open_problems}



\subsection{Temporal Splines}
\label{sec:open_problems:temporal_splines}

\subsubsection{Knot Selection}
\label{sec:open_problems:temporal_splines:knot_selection}

The optimal selection of knots remains \revision{an} important unresolved problems of the spline method.
The simplest and most common approach is to use uniform splines, tuning the frequency according to the application.
While uniform splines benefit from a fixed blending matrix that needs to be computed only once, they implicitly assume that excitation of the state evolution (e.g., robot motion) remains constant.
Hence, uniform knots can lead to underfitting or overfitting of the trajectory in settings with varying dynamics~\cite{hug2020hyperslam}.
Furthermore, a control point is only well-constrained if there are sufficient measurements on the associated segments, meaning that optimization can be poorly posed if the knot frequency is too high relative to the measurement frequency.

Some works associate knots with the measurement~\cite{bibby2010hybrid,yang2021asynchronous}, with \citet{bibby2010hybrid} selectively removing them where acceleration and jerk are sufficiently similar.
Others~\cite{vandeportaele2017pose,lang2023coco} increase the knot frequency according to the maximum magnitudes of acceleration and angular velocity (from \acs{IMU} measurements).
Another suggestion is to iteratively add knots until the mean visual reprojection error is sufficiently low~\cite{vandeportaele2017pose} or to increase knots until the reprojection residuals within segments agree with an expected value~\cite{oth2013rolling}.
\citet{anderson2014hierarchical} use this idea to decide when to increase the active resolution level in their hierarchical formulation, based on work by \citet{gortler1995hierarchical} who apply wavelet detail functions over a base B-spline signal function.

An alternative approach is \textit{spline error weighting}~\cite{ovren2018spline,ovren2019trajectory}, which recognizes that existing methods neglect approximation error arising from model mismatch.
By formulating spline fitting in the frequency domain, they devise a method to capture model mismatch and measurement error by setting the residual covariance based on the measurement noise and variance of the error.
They further use this frequency analysis to propose a method for selecting a suitable uniform knot frequency based on the fraction of signal energy to be retained.

Several \revision{other} schemes have been offered.
\citet{dube2016non} propose several distributions for the knots within the sliding window of the optimization.
These include an exponential distribution, a frequency analysis distribution based on average correction signal power, and a uniform distribution, potentially in combination.
\citet{hug2020hyperslam} discuss optimization of the knots themselves and derive Jacobians for them.
However, this idea has not yet been pursued experimentally.
Regularizing motion terms~\cite{furgale2015continuous,persson2021practical,johnson2024continuous} may avoid the issue of under-constrained control points; however, it may add assumptions conflicting with the implicit spline constraints.

\subsubsection{Certifiability \& Recovery Guarantees}
\label{sec:open_problems:temporal_splines:certifiability_and_recovery_guarantees}

Important open research questions arise because the optimization methods commonly used for estimation may converge to bad local minima or fail to converge altogether.
Indeed, depending on the measurement modalities used, the residuals of the factor graph (i.e., optimization cost terms) may be non-convex.
Particularly, if states are confined to manifold spaces, this may lead to non-convex constraints and exacerbate the problem.
Multiple recent works~\cite{holmes2023efficient,goudar2024optimal,yang2020teaser,rosen2019se} show that non-convexity may give rise to bad local minima in which local solvers such as \ac{GN} can get stuck.
\revision{In the face of these convergence issues, the global minimum can still be recovered from a tight but more costly convex relaxation (e.g., for initialization~\cite{goudar2024optimal}).}

A related question is that of the uniqueness of the solution.
Because splines promote smoothness, fewer measurements may be required to obtain unique state estimates.
However, exactly how many are necessary remains an open problem, and failure to meet this may result in ambiguities.
Recovery guarantees have been derived for polynomial or sinusoidal basis functions with noiseless range measurements~\cite{pacholska2020relax}, but extending these to more general cases remains open.

\subsubsection{\revision{Complex Processes}}
\label{sec:open_problems:temporal_splines:complex_processes}

\revision{The piecewise-polynomial interpolation of splines may be problematic in estimating high-frequency or hybrid dynamic systems, such as those that may occur in the presence of vibration or impacts.
Rather than increase knot frequency or spline order, which raises the computational load, \citet{li2024hcto} propose using the \acs{RMSE} of the accelerometer measurements to adaptively select between using \ac{IMU} measurements directly or preintegrating them as a form of low-pass filter.
Additionally, they apply constant-velocity factors to the roll and pitch when the acceleration is sufficiently low.
Resilience to non-polynomial and hybrid dynamics remains a challenging problem for the spline method.
}

\subsubsection{Incremental Optimization}
\label{sec:open_problems:temporal_splines:incremental_optimization}

Sliding-window optimization \revision{can} facilitate real-time operation~\cite{zlot2013efficient,zlot2014efficient,dube2016non,lowe2018complementary,quenzel2021real,mo2022continuous,tirado2022jacobian,lang2022ctrl,lv2023continuous,lang2023coco,nguyen2024eigen,lv2024cta}.
Other incremental schemes, such as those based on the Bayes tree~\cite{kaess2012isam2}, have not been explored.
Various graphical models have been presented for splines~\cite{hug2022continuous,mo2022continuous,tirado2022jacobian}, however only very recent works~\cite{lang2022ctrl,lv2023continuous,lang2023coco,lv2024cta} have made serious efforts to represent spline estimation with factor graphs and use these to design marginalization strategies.
Leveraging these to incorporate incremental optimizers like \acs{ISAM2}~\cite{kaess2012isam2} could be explored in future work.

\subsubsection{Initialization}
\label{sec:open_problems:temporal_splines:initialization}

Initialization of control points is rarely discussed in the literature.
However, appropriate initialization certainly affects convergence time and optimality.
Simple strategies for initializing new control points include using the value of the previous control point, or \ac{IMU} dead reckoning.
However, the state estimate may be far from the control point, especially for highly dynamic motion and high-order splines, and no inverse interpolation function exists to compute control points from the state.
One strategy adopted by \citet{huang2021b} was to use a spline approximation algorithm from \citet{piegl2012nurbs} for initialization.
\citet{li2024continuous} devise a sensor-specific coarse-to-fine procedure to initialize control points, biases, and extrinsic calibration parameters.
\citet{kang1999cubic} offer initialization equations for different parameterizations of orientation splines based on boundary conditions.

\subsubsection{Covariance Interpolation}
\label{sec:open_problems:temporal_splines:covariance_interpolation}

It is often desirable to infer the uncertainty of the state at the interpolation time from the uncertainty of the control points.
\revision{To a limited extent, covariance interpolation for vector-valued splines has been demonstrated~\cite[Figure 6]{furgale2015continuous}\cite{bibby2010hybrid}.}
\citet{bibby2010hybrid} provide a derivation for their non-uniform cubic splines, applied to dynamic \ac{SLAM} in 2D, using the Jacobian of the spline interpolation function. 
State covariance is also estimated by the probabilistic filter proposed by \citet{li2023embedding}.
However, no works provide a covariance interpolation formulation for Lie group splines.

\subsection{\acfp{TGP}}
\label{sec:open_problems:temporal_gaussian_processes}

\subsubsection{State Time Selection}
\label{sec:open_problems:temporal_gaussian_processes:state_time_selection}

The \ac{TGP} method provides flexibility in selecting state times.
This is analogous to the spline knot selection problem.
However, several differences in the methodology must be considered.
First, the process state variables are always constrained through \ac{GP} prior factors, meaning that \acp{TGP} are less sensitive to sensor dropout and do not overfit like splines.
Additionally, while measurements can be incorporated at any time, there are computational advantages of using non-interpolated factors, achieved by setting the state time to the measurement time.
Finally, when using local \acs{LTV}-\acsp{SDE}, the approximation error of the linearization is more significant when the changes between states are more considerable.
Thus, intuitively, the temporal density of states should be higher when the motion is more dynamic and less when the state changes slowly.
Regardless, the approaches proposed for spline knot selection may also apply to \acp{TGP}.

A straightforward approach is to set the state times as the measurement times of low-rate sensors, exploiting the efficiency of non-interpolated factors.
This is more advantageous when many factors are associated with a single measurement time, such as landmark measurements in global shutter visual \ac{SLAM}.
Adapting the state times based on \ac{IMU} measurements or residual errors as a surrogate metric for motion change would also be possible.
For more theoretical approaches, increasing knots until residual errors match expected values~\cite{oth2013rolling} could be applied to state times.
Alternatively, the state times - usually fixed dependent variables - could be added as optimization variables, as discussed for knots~\cite{hug2020hyperslam}.
Furthermore, residuals could be weighted analogously to spline error weighting~\cite{ovren2018spline,ovren2019trajectory} to account for the model approximation error uncaptured by the \acs{LTV}-\acs{SDE} prior.

\subsubsection{Certifiability \& Recovery Guarantees}
\label{sec:open_problems:temporal_gaussian_processes:certifiability_and_recovery_guarantees}

As for splines, common measurement models and state constraints may yield non-convex optimization problems that challenge convergence and global optimality guarantees for local solvers such as \ac{GN}.
A series of works on (\acs{DT}) globally optimal state estimation~\cite{carlone2015planar,rosen2019se,yang2020teaser} shows that many \ac{MAP} problems can be formulated as \acp{QCQP}, allowing for tight (convex) \ac{SDP} relaxations.
The latter can be used to \textit{globally} solve or certify \ac{CT} estimation problems.

In contrast with the spline method, \ac{GP} motion priors can be effortlessly integrated with this paradigm, as shown recently for the examples of \ac{CT} range-only localization~\cite{dumbgen2022safe} and pose-graph optimization~\cite{barfoot2024certifiably}.
Indeed, to incorporate a \ac{GP} motion prior, the state is augmented with the necessary number of temporal derivatives, after which the motion prior regularization always takes the form of additional inter-state quadratic cost terms, which can be readily included in the \ac{QCQP} formulation.
However, augmenting the state and regularizing the problem in this way may affect the tightness of the \ac{SDP} relaxations of the \ac{QCQP}, which, in turn, can break certifiability.
For pose-graph optimization in particular, a significant number of so-called \textit{redundant constraints} are required for tightness at reasonable noise levels when using the \ac{WNOA} prior~\cite[Equation (57)]{barfoot2024certifiably}, which breaks typical constraint qualifications and significantly increases the computational complexity of the used \ac{SDP} solvers.
Measures to increase and (a-priori) understand tightness and to speed up \ac{SDP} solvers are crucial for advancing the practicality of these methods.

As for the required number of measurements, \acp{TGP} behave differently than splines because of the Bayesian nature of the approach.
For example, in the absence of measurements, the method will return a trajectory according to the motion prior.
This allows the method to interpolate between states even when no measurements are acquired, while the uncertainty of such interpolations will grow.

\subsubsection{Complex \revision{Processes}}
\label{sec:open_problems:temporal_gaussian_processes:complex_processes}

\citet{tang2019white} show that selecting a process model prior that does not represent state evolution leads to biased estimation.
Thus, it is clear that a prior that most appropriately reflects the state evolution should be chosen.
\acp{TGP} inherently allows for complex process dynamics and control inputs to regularize the trajectory and guide interpolation.
\revision{However, only relatively simple models, such as the \ac{WNOA} or \ac{WNOJ} priors, have been examined thus far.
While efficient to compute and require tuning only a small number of hyperparameters, they do not accurately approximate the dynamics of all systems.
Systems with hybrid dynamics or impacts (e.g., legged robotics) may be particularly problematic since impulse-like accelerations induce perceived velocity discontinuities, violating the assumptions of existing priors.
Further research is needed into the feasibility and tractability of system-specific priors and control input utilization.
}

Some recent works have used learning in the modeling of system dynamics for fast online use~\cite{cioffi2023learned,cioffi2023hdvio}, and it would be reasonable to consider this idea in the computation or augmentation of the prior.
A principled method for tuning hyperparameters (e.g., $\mathbf{Q}_C$) is the maximization of the marginal log-likelihood using a training data set~\cite{anderson2015batch} (over hand-tuning~\cite{tang2019white,mukadam2018continuous,zhang2024gnss} or fitting to data~\cite{anderson2015full}), such as with the Singer prior~\cite{wong2020data}.
While this typically requires ground truth data, the \ac{EM} parameter learning framework extending \ac{ESGVI}~\cite{barfoot2020exactly} proposed by \citet{wong2020variational} demonstrates that hyperparameters can be learned from the original noisy measurement data.
The Singer prior~\cite{wong2020data} is an example of an alternative approach to process \revision{modeling}, whereby an \acs{SDE} is derived from factorization of the spectral density of a \ac{GP} prior, such as a kernel from the Mat\'ern family~\cite{hartikainen2010kalman,hartikainen2012sequential}.
It remains to be seen what methods could be effectively employed to learn reasonable \ac{GP} priors for complex processes.

\subsection{\revision{Evaluation and Benchmarking}}
\label{sec:open_problems:evaluation_and_benchmarking}

\revision{
Public benchmarks and appropriate metrics must be established with which \ac{CT} algorithms can be compared and evaluated.
While many works provide comparisons on public benchmarks to other hand-picked methods~\cite{lv2023continuous,nguyen2024eigen,zheng2024trajlo,zheng2024trajlio,lang2023coco,li2021towards,jung2023asynchronous}, they are frequently inconsistent, adding ambiguity.
Since \ac{CT} methods can fuse high-rate asynchronous measurements, benchmarks must consider how to provide sensor data and ground truth references.
For example, MCD~\cite{nguyen2024mcd} recently leveraged survey-grade maps to optimize a ``ground truth'' B-spline from \acs{LIDAR}-inertial data.
Existing algorithms are usually compared with aggregated absolute or relative trajectory errors, which are insufficient metrics for understanding their complete behavior.
\citet{zhang2019rethinking} demonstrate the capacity for temporal \acp{GP} with Gaussian kernels to represent ground truth trajectories and propose generalized absolute and relative errors for principled trajectory evaluation.
}

\section{Applications to other Domains}
\label{sec:applications_to_other_domains}

\subsection{Temporal Splines}
\label{sec:applications_to_other_domains:temporal_splines}

Splines have been extensively used for planning in robotics, including wheeled robots~\cite{elbanhawi2015randomized}, legged locomotion~\cite{bellicoso2018dynamic},  wheel-legged locomotion~\cite{bjelonic2020rolling}, and manipulation~\cite{chand1985line,piazzi2000global}.
However, due to their smoothness guarantees, splines are especially appealing for \emph{differentially flat systems}, such as \acp{UAV}~\cite{van1998real,mellinger2011minimum}.
In a differentially flat system, the states and the inputs can be written as algebraic functions of some variables (called \emph{flat outputs}) and their derivatives. 
The continuity of splines can guarantee that the \acs{UAV}'s trajectory is dynamically feasible.

Moreover, splines whose basis functions form a partition of unity (i.e., \revision{nonnegative, summing to one}) exhibit the \emph{convex hull} property, which guarantees that each spline segment is wholly contained within the convex hull of its control points.
This is leveraged by many trajectory planning works~\cite{sahingoz2014generation,ding2018trajectory,zhou2019robust,tordesillas2019faster,tordesillas2020mader,tordesillas2021panther,preiss2017trajectory}, since it substantially simplifies obstacle avoidance constraints~\cite{liu2017planning}.
B-splines are the typical choice of splines for trajectory planning, as they can guarantee smoothness and high continuity by construction without imposing explicit constraints.
However, the convex hull of the B-Spline control points is typically much larger than the segment itself.
To reduce conservativeness, some works~\cite{tang2019real} use B-spline control points as decision variables in trajectory planning, but then, the B\'ezier control points of each interval of this B-spline are used in the obstacle avoidance constraints.
The MINVO control points~\cite{tordesillas2022minvo,herron1989polynomial} allow for even tighter enclosures and have been leveraged for obstacle avoidance~\cite{herron1989polynomial}.


To a limited degree, shape estimation for \emph{continuum} or flexible robots has also been found to be amenable to spline and \ac{TBF} theory since they are also one-dimensional estimation problems, parameterized by arc length instead of time.
For example, weighted combinations of basis functions~\cite{lobaton2013continuous,kim2014optimizing} and B\'ezier curves~\cite{song2015electromagnetic} have been explored.

Splines have also been leveraged to represent specific environment features.
For instance, Catmull-Rom splines are used by \citet{qiao2023online} to parameterize the road lanes.
B-splines are used by \citet{rodrigues2018b} to represent unstructured 2D environments more compactly, and in other works~\cite{pedraza2009extending,liu2009new,liu2010towards} to represent the sensor data in their state estimation frameworks.
Spline theory has also been utilized for perception outside the state estimation problems discussed in this work, such as the spatio-temporal representation of event camera feature tracks with B\'ezier curves~\cite{seok2020robust}.

In machine learning, apart from traditional regression~\cite{marsh2001spline, sangalli2013spatial}, splines have also recently found applications in neural networks~\cite{fakhoury2022exsplinet, zhu2020gssnn, durkan2019neural}, including recently \acp{KAN}~\cite{liu2024kan}, proposed as a promising alternative to the traditional \ac{MLP}. Moreover, \citet{yang2023iplanner} and \citet{roth2024viplanner} use splines as the output of their end-to-end learned neural network-based local planner, trained imperatively through Bilevel optimization. 

\subsection{\acfp{TGP}}
\label{sec:applications_to_other_domains:temporal_gaussian_processes}

\acp{GP} have been deployed for a large number of other tasks in robotics.
\acp{GP} with Gaussian kernels have \revision{often} been used in Bayesian filters, such as for replacing or enhancing the parametric prediction and observation models~\cite{ko2007gp,ko2009gp,ko2011learning}.
The use of \acp{GP} as surrogate observation models has been explored for signal-strength (e.g., WiFi, magnetic flux) state estimation applications~\cite{hahnel2006gaussian,ferris2007wifi,brooks2006gaussian,mcdonald2023global}, and as dynamics prediction models within the control community~\cite{ko2007gaussian_2,nguyen2008local,deisenroth2013gaussian,boedecker2014approximate,cao2024computation}.
In informative path planning applications, \acp{GP} have been used for representing scalar environmental fields, like temperature, salinity, methane concentration, magnetic field intensity, and terrain height~\cite{viseras2016decentralized,allamraju2017communication,ma2017informative,mishra2018online,jang2020multi,wakulicz2022informative}.
\acp{GP} have also seen extensive use in mapping, including occupancy maps~\cite{o2009contextual,gan20093d,gan20093d_2,o2012gaussian,kim2012building,kim2013continuous,wang2016fast,o2016gaussian,hata2017monte,jadidi2017warped,yuan2018fast,ghaffari2018gaussian}, implicit surfaces~\cite{williams2006gaussian_2,dragiev2011gaussian,gerardo2013laser,gerardo2014robust,martens2016geometric,caccamo2016active,lee2019online,wu2020skeleton,wu2021faithful,liu2021active,ivan2022online,wu2023log}, their combination~\cite{kim2015gpmap}, gradient maps~\cite{le2020gaussian_2,giubilato2020gpgm,giubilato2021gpgm,giubilato2022robust} and distance fields~\cite{wu2021faithful,wu2023log,wu2023pseudo,le2023continuous_2,le2023accurate,wu2024vdb}.
They have also been used as a representation for local perception and exploration uncertainty~\cite{ali2023autonomous}.
However, these works did not use \acs{LTV}-\acs{SDE} priors to obtain exact sparsity in \ac{GP} regression.\footnote{Although, some works~\cite{kim2015gpmap} attain sparsity in the covariance matrix by using a piecewise kernel that is zero beyond a certain distance~\cite{melkumyan2009sparse}.
} 

\citet{mukadam2016gaussian} were the first to apply exactly sparse \acp{TGP} to the problem of motion planning with the \acs{GPMP} algorithm, in particular for a 7-\acs{DOF} robotic arm, addressing the large number of states needed by discrete-time planners.
This was extended to factor graph optimization, in GPMP2~\cite{dong2016motion} (later used by \citet{maric2018singularity}), GPMP-GRAPH~\cite{huang2017motion}, and iGPMP2~\cite{dong2016motion,mukadam2018continuous}, before unification with state estimation in \acs{STEAP}~\cite{mukadam2017simultaneous,mukadam2019steap}, and learning from demonstration in \acs{CLAMP}~\cite{rana2018towards}.
\acs{CLAMP} is unique from other works in that the transition matrix $\mathbf{\Phi}(t_{i+1}, t_i)$ and bias $\mathbf{v}(t_{i+1}, t_i)$ are learned for particular tasks from trajectory demonstrations with linear ridge regression.
GPMP2 was also extended to dGPMP2~\cite{bhardwaj2020differentiable}, which through a self-supervised end-to-end training framework, allowed the learning of factor covariance parameters, such as $\mathbf{Q}_C$, an idea which could be applied to state estimation in future work.
Inspired by these works, \citet{cheng2022real} adopt an exactly sparse \ac{GP} for path planning in an autonomous driving context, parameterizing their \ac{WNOJ} prior by arc length instead of time.
\citet{lilge2022continuum,lilge2024state} similarly parameterize by arc length to model the shape of a continuum robot using a white-noise-on-strain-rate (analogous to \ac{WNOA}) \ac{LTV}-\ac{SDE} prior derived from the Cosserat rod model, with state consisting of $SE(3)$ pose and $\mathbb{R}^6$ generalized strain.
\revision{They recently explore including both temporal and spatial domains jointly~\cite{teetaert2024space}.}

\section{Conclusion}
\label{sec:conclusion}

This work examined the literature and methods of continuous-time state estimation for robotics.
Interpolation and integration techniques remain popular \ac{CT} methods, especially linear interpolation due to its simplicity and low computation cost.
This work has focused more heavily on temporal splines and Gaussian processes, which can represent complex process dynamics while remaining computationally competitive.
By modeling state evolution explicitly, rather than capturing `snapshots' at measurement times, \ac{CT} methods offer significant advantages over \ac{DT} estimation, both for the performance, flexibility, and generalizability of the estimator itself and for downstream planning or control algorithms.
By providing introductory theory, a comprehensive survey, and a discussion of open problems and applications in other domains, this work aims to support future \ac{CT} research, which will become increasingly essential to the field as the number, variety, and complexity of sensors grow in robotics applications.

\begin{table}[ht]
    \centering
    \caption{Variable definitions. If omitted, $n$ is assumed to be 3 (3D).}
    \vspace{-3.5mm}
    \resizebox{\linewidth}{!}{
    \def\arraystretch{0.9}
    \rowcolors{2}{gray!8}{white}
    \begin{tabular}{|l|l|l|}
        \hline
        \textbf{Variable} & \textbf{State Representation(s)} & \textbf{Description} \\
        \hline
        ${}^n\mathbf{p}$ & $\mathbb{R}^n$ & $n$-Dimensional Position \\
        ${}^n\dot{\mathbf{p}}$ & $\mathbb{R}^n$ & $n$-Dimensional Linear Velocity \\
        ${}^n\ddot{\mathbf{p}}$ & $\mathbb{R}^n$ & $n$-Dimensional Linear Acceleration \\
        ${}^n\mathbf{r}$ & $SO(n), \mathbb{R}^{n(n-1)/2}, SU(2)$ & $n$-Dimensional Orientation \\
        ${}^n\dot{\mathbf{r}}$ & $\mathbb{R}^{n(n-1)/2}$ & $n$-Dimensional Angular Velocity \\
        ${}^n\ddot{\mathbf{r}}$ & $\mathbb{R}^{n(n-1)/2}$ & $n$-Dimensional Angular Acceleration \\
        ${}^n\mathbf{T}$ & $SE(n), {}^n\mathbf{r} \times {}^n\mathbf{p}$ & $n$-Dimensional Pose \\
        ${}^n\dot{\mathbf{T}}$ & $\mathbb{R}^{n(n+1)/2}$ & $n$-Dimensional Pose Velocity \\
        ${}^n\ddot{\mathbf{T}}$ & $\mathbb{R}^{n(n+1)/2}$ & $n$-Dimensional Pose Acceleration \\
        \hline
        $\mathcal{S} = \{\mathcal{S}_E, \mathcal{S}_I\}$ & - & Sensor Calibration Variables\\
        $\mathcal{S}_I$ & - & Sensor Intrinsics\\
        $\mathbf{K}$ & - & Camera Projection Intrinsics \\
        $t_\text{\acs{RS}}$ & $\mathbb{R}$ & Rolling Shutter Line Delay \\
        $t_e$ & $\mathbb{R}$ & Camera Exposure Time \\
        $\bm{r}_C$ & $\mathbb{R}^n$ & Camera Response Parameters \\
        $\bm{d}_C$ & $\mathbb{R}^n$ & Camera Distortion Parameters \\
        $b_r$ & $\mathbb{R}$ & Range Bias \\
        $s_r$ & $\mathbb{R}$ & Range Scale \\
        $\mathbf{m}_r$ & - & Range Frame Misalignment Parameters \\
        $\mathbf{b}_g$ & $\mathbb{R}^3$ & Gyroscope Bias \\
        $\mathbf{S}_g$ & $\mathbb{R}^{3 \times 3}~\text{(diagonal)}$ & Gyroscope Scale \\
        $\mathbf{M}_g$ & $\mathbb{R}^{3 \times 3}~\text{(triangular)}$ & Gyroscope Axis Misalignment \\
        $\mathbf{G}_g$ & $\mathbb{R}^{3\times 3}$ & Gyroscope G-Sensitivity\\
        $\mathbf{b}_a$ & $\mathbb{R}^3$ & Accelerometer Bias\\
        $\mathbf{S}_a$ & $\mathbb{R}^{3 \times 3}~\text{(diagonal)}$ & Accelerometer Scale \\
        $\mathbf{M}_a$ & $\mathbb{R}^{3 \times 3}~\text{(triangular)}$ & Accelerometer Axis Misalignment \\
        $b_{\text{\acs{GNSS}, t}}$ & $ \mathbb{R}$ & \acs{GNSS} Receiver Clock Bias \\
        $b_{\text{\acs{GNSS}, v}}$ & $ \mathbb{R}$ & \acs{GNSS} Receiver Velocity Bias \\
        $n_{\text{\acs{GNSS}}}$ & $\mathbb{N}^0$ & Number of Satellites \\
        \hline
        $\mathcal{S}_E = \{\mathbf{T}_E, \mathbf{t}_E\}$ & - & Sensor Extrinsics \\
        $\mathbf{T}_E$ & $(\mathbf{T})^n$ & Inter-sensor Relative Poses \\
        $\mathbf{p}_{ia}$ & $\mathbb{R}^3$ & \acs{IMU}-Accelerometer Translation \\
        $\mathbf{r}_{ga}$ & $\mathbf{r}$ & Gyroscope-Accelerometer Rotation \\
        $\mathbf{t}_E$ & $\mathbb{R}^n$ & Inter-sensor Time Offsets \\
        \hline
        $\mathbf{g}$ & $\mathbb{R}^3$ & Gravity Vector\\
        $\mathbf{r}_\mathbf{g}$ & $\mathbb{R}^2, \mathbf{r}$ & Gravity Orientation\\
        $s$ & $\mathbb{R}$ & Map Scale\\
        ${}^n\mathcal{M}$ & - & Map in $n$-Dimensional Space\\
        ${}^n\mathcal{M}_S$ & - & Surfel Map in $n$-Dimensional Space\\
        ${}^n\mathcal{M}_O$ & - & Occupancy Map in $n$-Dimensional Space\\
        ${}^n\mathbf{L}$ & $\mathbb{R}^{n \times L}$ & Point Landmarks in $n$-Dimensional Space\\
        $\mathbf{L}_\pi$ & $\mathbb{R}^{4 \times L}$ & Plane Landmarks in 3D Space\\
        $\mathbf{p}_d$ & $\mathbb{R}^p$ & Pixel/Landmark Depths/Distances\\
        $\bm{\rho}_d$ & $\mathbb{R}^p$ & Pixel/Landmark Inverse Depths/Distances\\
        $\mathbf{p}_b$ & $\mathbb{R}^{n \times p}$ & Pixel/Landmark Bearing Vectors\\
        $\mathbf{m}_o$ & $\mathbb{R}^n$ & Material Optical Parameters \\
        \revision{$\mathbf{T}_D$/$\dot{\mathbf{T}}_D$/$\ddot{\mathbf{T}}_D$} & \revision{$\mathbf{T}$/$\dot{\mathbf{T}}$/$\ddot{\mathbf{T}}$} & \revision{Dynamic Object Pose/Velocity/Acceleration} \\
        \revision{$\mathbf{p}_D$/$\dot{\mathbf{p}}_D$/$\ddot{\mathbf{p}}_D$} & \revision{$\mathbf{p}$/$\dot{\mathbf{p}}$/$\ddot{\mathbf{p}}$} & \revision{Dynamic Object Position/Velocity/Acceleration} \\
        \hline
    \end{tabular}
    }
    \vspace{-7mm}
    \label{tab:variable_definitions}
\end{table}

{
\renewcommand*{\bibfont}{\scriptsize} 
\linespread{0.863} 
\bibliographystyle{IEEEtranN}
\bibliography{references_custom_abrv} 
}

\end{document}